\documentclass[Afour,sageh,times]{sagej}

\usepackage{moreverb,url}

\usepackage[colorlinks,bookmarksopen,bookmarksnumbered,citecolor=red,urlcolor=red]{hyperref}

\usepackage[utf8]{inputenc}
\usepackage[english]{babel}

\usepackage{algorithm}
\usepackage{algpseudocode}
\usepackage{float}
\usepackage{xcolor}
\usepackage{caption}
\usepackage{subcaption}
\usepackage{tikz}
\usepackage{amsmath}
\usepackage{wrapfig}
\usepackage{lipsum}
\usepackage{moreverb,url}
\usepackage{xcolor}
\usepackage[normalem]{ulem}
\usepackage[colorlinks,bookmarksopen,bookmarksnumbered,urlcolor=red]{hyperref}

\newcommand\BibTeX{{\rmfamily B\kern-.05em \textsc{i\kern-.025em b}\kern-.08em
T\kern-.1667em\lower.7ex\hbox{E}\kern-.125emX}}

\usetikzlibrary{matrix,backgrounds}
\usepackage{natbib}
\def\volumeyear{2022}

\begin{document}

\runninghead{Aguasvivas Manzano et. al.}

\title{Toward smart composites: small-scale, untethered prediction and control for soft sensor/actuator systems}

\author{Sarah {Aguasvivas Manzano}\affilnum{1}, Vani Sundaram \affilnum{2}, Artemis Xu \affilnum{3}, Khoi Ly \affilnum{2,3}, Mark Rentschler\affilnum{2}, Robert {Shepherd} \affilnum{3} and Nikolaus {Correll}\affilnum{1}}

%% Vani: Paper old all stuff showing here is old paper 

\affiliation{\affilnum{1}Department of Computer Science, University of Colorado, Boulder, CO, USA \\
\affilnum{2}Paul M. Rady Department of Mechanical Engineering, University of Colorado, Boulder, CO, USA \\
\affilnum{3}Sibley School of Mechanical and Aerospace Engineering, Cornell University, Ithaca, NY, USA}

\corrauth{Sarah Aguasvivas Manzano,
Department of Computer Science,
University of Colorado, Boulder, CO, USA}
\email{Sarah.AguasvivasManzano@colorado.edu}

\begin{abstract}

We present formulation and open-source tools to achieve in-material model predictive control of sensor/actuator systems using learned forward kinematics and on-device computation. Microcontroller units (MCUs) that compute the prediction and control task while colocated with the sensors and actuators enable in-material untethered behaviors. In this approach, small parameter size neural network models learn forward kinematics offline. Our open-source compiler, \emph{nn4mc}, generates code to offload these predictions onto MCUs. A Newton-Raphson solver then computes the control input in real time. We first benchmark this nonlinear control approach against a PID controller on a mass-spring-damper simulation. We then study experimental results on two experimental rigs with different sensing, actuation and computational hardware: a tendon-based platform with embedded \emph{LightLace} sensors and a HASEL-based platform with magnetic sensors. Experimental results indicate effective high-bandwidth tracking of reference paths ($\geq$ 120 Hz) with a small memory footprint ($\leq$ 6.4\% of flash memory). The measured path following error does not exceed 2mm in the tendon-based platform. The simulated path following error does not exceed 1mm in the HASEL-based platform. The mean power consumption of this approach in an ARM Cortex-M4f device is 45.4 mW. This control approach is also compatible with Tensorflow Lite models and equivalent on-device code. In-material intelligence enables a new class of composites that infuse autonomy into structures and systems with refined artificial proprioception.

\end{abstract}


\maketitle

\begin{figure}[H]
    \centering
    \begin{subfigure}[t]{\columnwidth}
    \centering
    \begin{tikzpicture}
    \begin{scope}
    \node (image) at (0, 0)
    {\includegraphics[width=0.86\columnwidth, trim={0 0 0 0},clip]{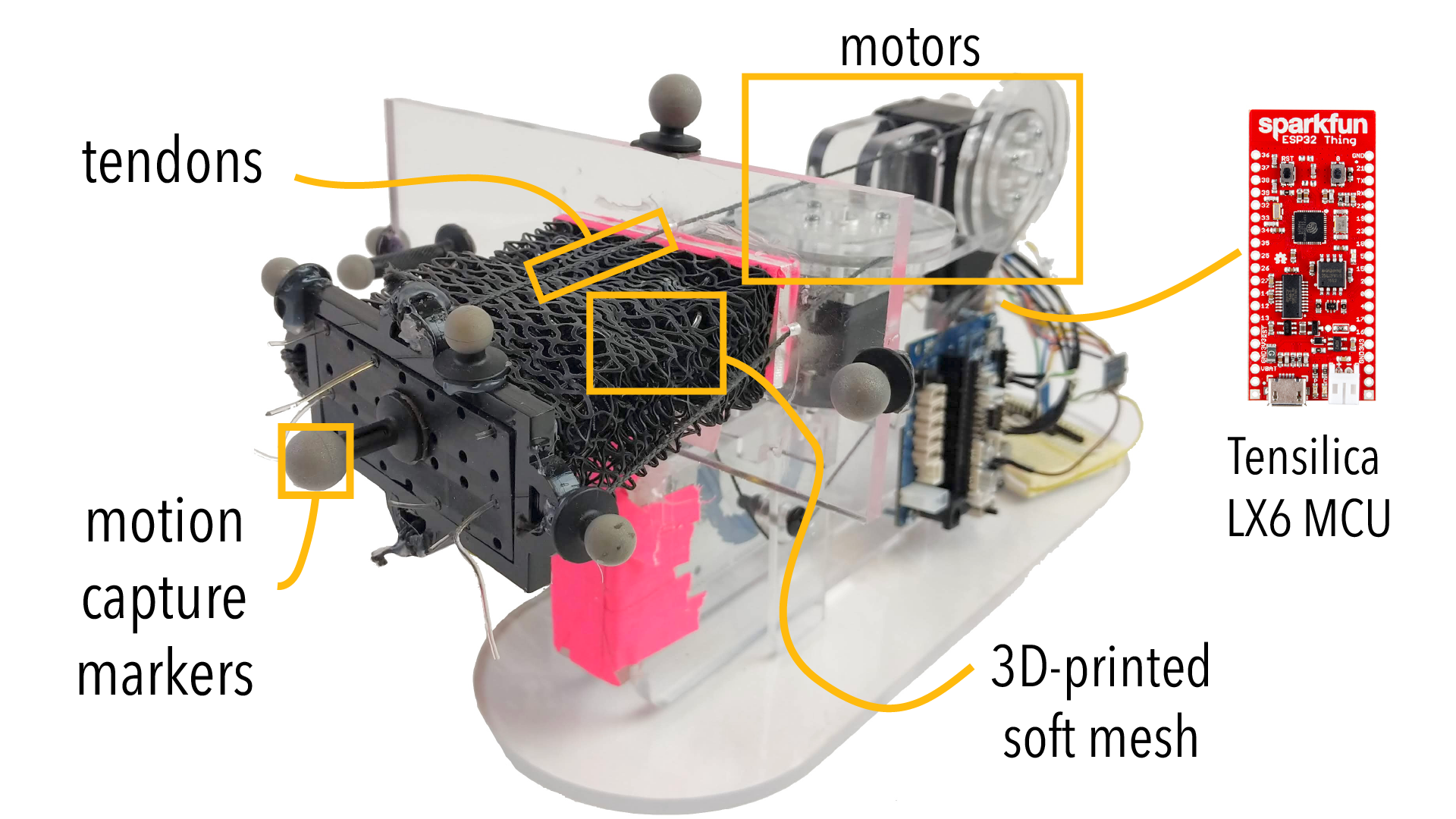}};
    \end{scope}
    \node (letter) at (-4, 1.6){\textbf{(a)}};
   \end{tikzpicture}
   \label{fig:test}
   \end{subfigure}
    \begin{subfigure}[t]{\columnwidth}
    \centering
    \begin{tikzpicture}
    \begin{scope}
    \node (image) at (0, 0)
    {\includegraphics[width=\columnwidth,trim={0 0 0 1.5cm},clip ]{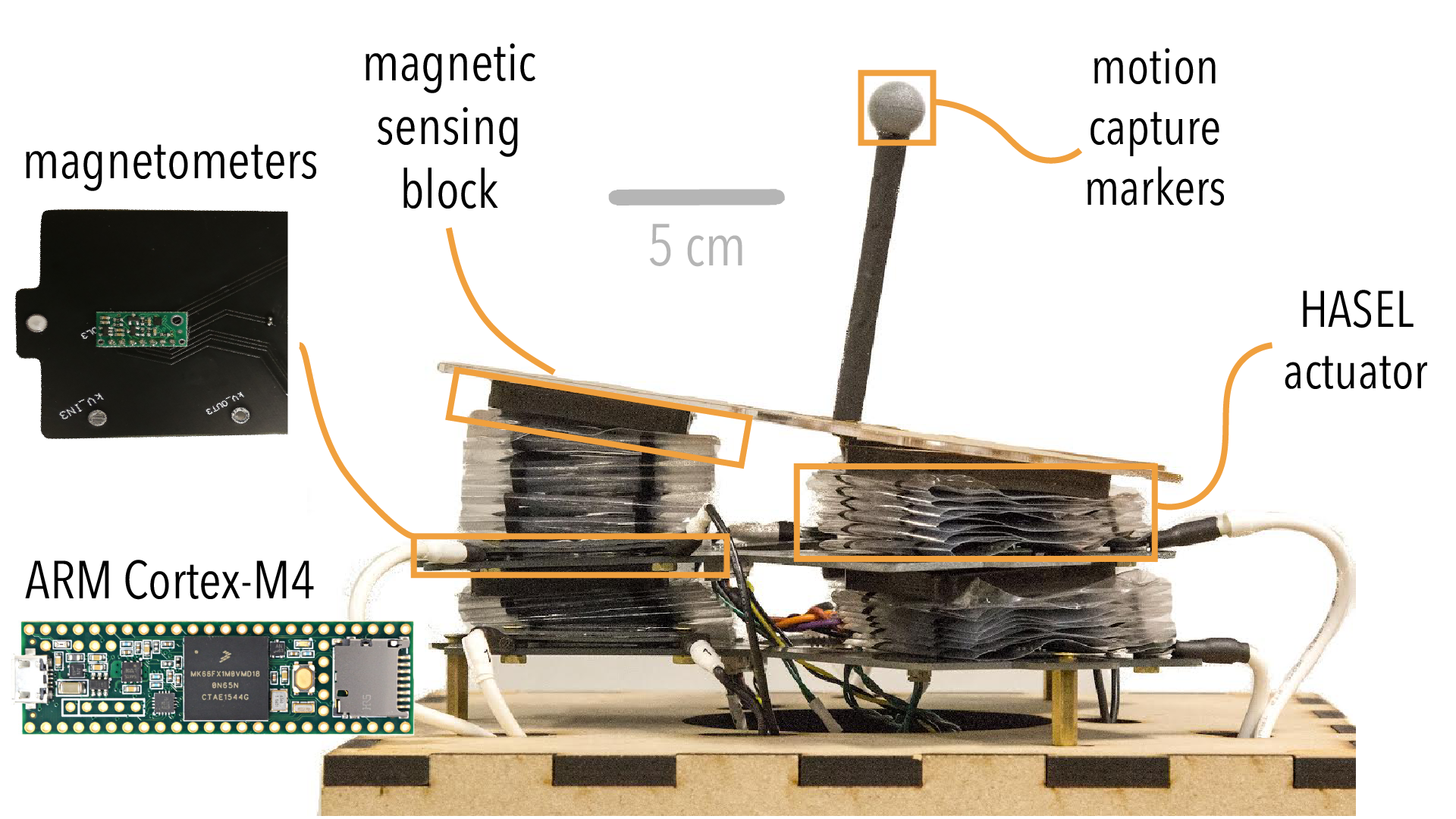}};
    \end{scope}
    \node (letter) at (-4, 2.3){\textbf{(b)}};
   \end{tikzpicture}
   \end{subfigure}

  \caption{
  Hardware platforms studied for on-board nonlinear neural-based control. (A) A tendon-based platform with embedded \textit{LightLace} sensors \citep{xu2019optical, manzano2020high} (B) A HASEL-based tilting platform with embedded magnetic sensors \citep{sundaram2022embedded}.
  }
  \label{fig:photographs}
\end{figure}

\section{Introduction}
\label{sec:introduction}

Soft sensors and actuators leverage materials and structures that exhibit large-scale deformation, high compliance, and rich multi-functionality, behaving in ways that are difficult to mimic with conventional, rigid mechatronic systems \citep{iida2011soft}. Through a rich set of alternative sensors and actuators, the field of soft robotics creates opportunities for multi-functional composite materials that tightly integrate sensing, actuation, computation and communication \citep{mcevoy2015materials}.

Yet, determining generalized models for soft sensors and actuators, and using them in closed-loop control are difficult tasks because soft sensors, actuators, and structural materials exhibit viscoelastic effects, such as hysteresis, stress relaxation, and creep. Additionally, the fabrication of soft actuators currently tends to include multiple handcrafted manufacturing stages, which can make it difficult to make sensors and actuators to behave predictably and precisely \citep{cho2009review}. 

Machine learning (ML) alleviates some of these challenges \citep{chin2020machine,kim2021review}; however, without proper optimizations, ML may also presents a computationally expensive approach due to its training resource needs, memory-hungry models, limited support for embedded systems, and computational efficiency \citep{dietterich2002machine}, which makes algorithms reliant on personal computers, clouds and external computers. While using powerful, external computing systems might be acceptable for tethered soft robotic systems, it makes deployment of soft sensors and actuators into composites at scale, with long battery life, elusive. 

Regarding ML for forward on-device inference, post-training quantization enables model compression at a loss for accuracy \citep{david2020tensorflow}, and co-training quantized models improve these accuracy losses \citep{lin2020mcunet}. In this work, we compile neural network inference into efficient C code and combine these models with efficient numerical methods to create an untethered, lightweight model-predictive controller in floating-point logic without losses in model accuracy at millimeter-scale task spaces. Throughout this work, we use the term ``untethered" interchangeably with the term ``computationally untethered," i.e., high-performance computation in the firmware of the embedded low-power compute units or units without being tethered to personal computers, clouds, joysticks, or any remote brains and using feedback from proprioception provided by networks of synthetic afferent sensor networks \citep{xu2019optical, sundaram2022embedded}.

\cite{manzano2020high} presents initial formulations and a Python prototype of the output tracking technique developed in this work for the tendon-based platform with embedded \textit{LightLace} sensors developed by \cite{xu2019optical}, we also study the robustness of this control method against mechanical disturbances. Figure \ref{fig:photographs}(a) displays the actuating, tendon-based, platform that uses feedback from this network of embedded \emph{LightLace} sensors. In this paper, we develop, test, and deploy fully functioning, open-source firmware controller code in C that is tractable in low-power MCUs. We analyze a simulated mass-spring-damper system, which approximates the expected kinematic motion of the class of soft actuators that this work studies in one dimension and isolates nonlinear effects coming from the composite materials. This paper further extends the details on the numerical computations and validates the system-independence of this approach using a system of hydraulically amplified self-healing electrostatic (HASEL) actuators \citep{acome2018hydraulically, mitchell2019easy}. Six HASEL actuators are arranged in a tilting platform, where the firmware not only collects sensor signals and actuates the HASELs, but also does all the computation for high-bandwidth neural-based feedback control. Figure \ref{fig:photographs}(b) pictures this platform containing magnetic sensors \citep{sundaram2022embedded} and HASEL actuators. Lastly, we test the claim of the tractability of this algorithm in low-power computers by measuring the power and modelling the energy consumption of this algorithm.

\subsection{Related Work}
\label{sec:related_work}

Besides several desired traits and behaviors in soft robots, such as compliance and motion biomimicry, hardware untethering is an essential requirement for many real-life applications that require low-power operations such as space \citep{ng2021untethered, lipson2014challenges, rus2015design} and deep-sea exploration \citep{giorgio2017hybrid, krieg2015design}, search and rescue missions \citep{mintchev2018soft, hawkes2017soft}, wearable technology \citep{matthies2021capglasses,teague2020wearable}, and smart composites \citep{mcevoy2015materials}. Equally as important, advanced software untethering guarantees the deployability of soft sensors and actuators in harsh environments that often require autonomy and advanced control at the low-level that is past the proportional-integral-differential (PID) control task due to a complex relationship between inputs, feedback and outputs in the system. State-of-the-art, untethered soft robots are difficult at the design and manufacturing stage and typically perform algorithmically simple tasks, such as crawling \citep{rich2018untethered, usevitch2020untethered}, which consist of hand-coded motions toward gait. In other cases, more advanced control is localized at segments of a smart composite such as shape-changing materials \citep{mcevoy2015thermoplastic} or smart skins \citep{hughes2015texture}, both of which have motivated the work presented here.

Model-free, learning-based programs, which have traditionally required powerful, tethered computers, outperform Jacobian-based approaches in controlling nonlinear soft systems \citep{giorelli2015neural} for a cable-driven soft arm with variable curvature. \cite{thuruthel2019soft} has shown that Long Short-Term Memory units (LSTM) are suitable for modeling the kinematic responses of soft actuators with embedded sensors in real-time while being robust against sensor drift. In \cite{hyatt2019model}, a 3.4-million-parameter neural network model that learns the forward dynamics of the system computes the predictions for model predictive control. In \cite{gillespie2018learning}, a fully connected neural network of three hidden layers with 200 nodes that represents a model discretized at 0.033s learns forward dynamics predictions for model predictive control. \cite{bruder2019modeling} achieved path tracking errors in the order of centimeters (average Euclidean error of 1.26 cm also using a learned approach. \cite{bern2020soft} used learning towards the inverse kinematic of a soft robot with three actuator inputs and achieved a very high-performance following of a prescribed path while using a neural network model and the MATLAB Deep Learning Toolbox \citep{matlabdeeplearning}.

``Neural Networks for Microcontrollers,'' or \textit{nn4mc}\footnote{https://nn4mc.com} is part of our compound efforts towards wireless prediction and control \citep{manzano2019embedded, manzano2018wireless}. Recent developments of \textit{nn4mc} include support for Gated Recurrent Units (GRU) layers and other functionalities. Other academic- and industry-led efforts towards deployed neural network intelligence include MCUNet \citep{lin2020mcunet}, which is a framework that jointly designs and trains quantized neural network architectures for increased accuracy. Other efforts towards machine learning on embedded devices include Tensorflow Lite \citep{david2020tensorflow, lai2018cmsis}, CMix-NN \citep{capotondi2020cmix}, MicroTVM \citep{chen2018tvm} and tinyML \citep{warden2019tinyml}. The advantage of \emph{nn4mc} compared to other available packages summarizes to universal compatibility with any microcontroller that interprets embedded C such as Arm-based  Teensy\footnote{https://www.pjrc.com/teensy/} or the Tensilica-based ESP32\footnote{https://www.espressif.com/en/products/modules}, both used in this work, while giving the researcher full access to the generated code.

Devising a controller that uses in-material neural network predictions allows for composite materials to deeply embed sensing, computation and actuation, thereby leading to ``materials that make robots smart'' \citep{hughes2019materials}, blurring the boundary between materials and computation \citep{mengucc2017will}. This new class of materials are capable of making decisions based on their own proprioception without the need for communication with external computers or external observers and take advantage of existing electronics for advanced computation.

\subsection{Contributions of this Work}
\label{sec:contributions}

We present a nonlinear, predictive controller capable of functioning at high bandwidth and low power. The controller is untethered from a personal computer through the use of small-sized recursive networks and online optimization using a Newton-Raphson solver \citep{soloway1996neural} in the MCU firmware. We aim for small memory footprint, thus low power consumption and efficient path tracking of an end-effector. This control method compares favorably with other state-of-the-art methods by increasing bandwidth, decreasing output tracking errors, increasing real-time smoothness, and increasing precision despite working for systems with highly nonlinear sensor readings actuator responses. An implementation of our framework is available open-source both in C and Python.\footnote{\href{https://github.com/sarahaguasvivas/nlsoft}{https://github.com/sarahaguasvivas/nlsoft}}

Accessible firmware engineering is critical for the continuous proliferation of soft robots and smart composite materials. To the authors' knowledge, no past efforts combine on-device neural network predictions with nonlinear onboard control in low-power MCU firmware of the soft robot to produce untethered nonlinear real-time feedback control with online optimization. \footnote{\href{https://youtu.be/DQ-uXf1tA3w}{Link to high-level project demo.}}

\section{Soft Actuator Behavior}
\label{sec:soft_actuators}
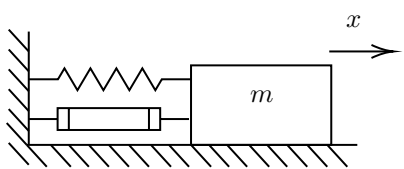
\begin{figure}[H]
\centering
\tikzset{every picture/.style={line width=0.75pt}} %set default line width to 0.75pt        

\begin{tikzpicture}[x=0.75pt,y=0.75pt,yscale=-1,xscale=1]

%Shape: Rectangle [id:dp038280114874792215] 
\draw   (99,26) -- (169,26) -- (169,66) -- (99,66) -- cycle ;
%Straight Lines [id:da8211608912190422] 
\draw    (18,9) -- (18,66) ;
%Straight Lines [id:da5998652041159636] 
\draw    (18,19) -- (8,8) ;
%Straight Lines [id:da15750271018595496] 
\draw    (18,29) -- (8,18) ;
%Straight Lines [id:da17146594605320065] 
\draw    (18,39) -- (8,28) ;
%Straight Lines [id:da9255618918599979] 
\draw    (18,49) -- (8,38) ;
%Straight Lines [id:da3523429368162123] 
\draw    (18,59) -- (8,48) ;
%Straight Lines [id:da008689208538507076] 
\draw    (118,77) -- (108,66) ;
%Straight Lines [id:da5171674459102145] 
\draw    (109,77) -- (99,66) ;
%Shape: Resistor [id:dp6443487669720831] 
\draw   (18,33) -- (32.58,33) -- (35.82,27.5) -- (42.3,38.5) -- (48.78,27.5) -- (55.26,38.5) -- (61.74,27.5) -- (68.22,38.5) -- (74.7,27.5) -- (81.18,38.5) -- (84.42,33) -- (99,33) ;
%Shape: Fuse [id:dp7938509581994262] 
\draw   (32.4,47.5) -- (83.6,47.5) -- (83.6,58.5) -- (32.4,58.5) -- (32.4,47.5) -- cycle (18,53) -- (32.4,53) (83.6,53) -- (98,53) (38,47.5) -- (38,58.5) (78,47.5) -- (78,58.5) ;
%Straight Lines [id:da38621040970014464] 
\draw    (168,19) -- (198,19) ;
\draw [shift={(200,19)}, rotate = 180] [color={rgb, 255:red, 0; green, 0; blue, 0 }  ][line width=0.75]    (10.93,-3.29) .. controls (6.95,-1.4) and (3.31,-0.3) .. (0,0) .. controls (3.31,0.3) and (6.95,1.4) .. (10.93,3.29)   ;
%Straight Lines [id:da07855319565832453] 
\draw    (17,66) -- (182,66) ;
%Straight Lines [id:da6054959926499044] 
\draw    (39,77) -- (29,66) ;
%Straight Lines [id:da590512149840394] 
\draw    (48,77) -- (38,66) ;
%Straight Lines [id:da33805840366107165] 
\draw    (58,77) -- (48,66) ;
%Straight Lines [id:da4965719407840701] 
\draw    (68,77) -- (58,66) ;
%Straight Lines [id:da19960855585920156] 
\draw    (78,77) -- (68,66) ;
%Straight Lines [id:da3249490678077547] 
\draw    (89,77) -- (79,66) ;
%Straight Lines [id:da5491208068290947] 
\draw    (98,77) -- (88,66) ;
%Straight Lines [id:da8256106757524262] 
\draw    (127,77) -- (117,66) ;
%Straight Lines [id:da34088971519278477] 
\draw    (138,77) -- (128,66) ;
%Straight Lines [id:da6895536788458982] 
\draw    (147,77) -- (137,66) ;
%Straight Lines [id:da25731524768532843] 
\draw    (157,77) -- (147,66) ;
%Straight Lines [id:da37367261238092553] 
\draw    (177,77) -- (167,66) ;
%Straight Lines [id:da2155890828164453] 
\draw    (167,77) -- (157,66) ;
%Straight Lines [id:da7412225750123358] 
\draw    (27,77) -- (17,66) ;
%Straight Lines [id:da9866960259085755] 
\draw    (17,66) -- (7,55) ;
%Straight Lines [id:da6538935578320342] 
\draw    (18,76) -- (8,65) ;

% Text Node
\draw (175,-0.6) node [anchor=north west][inner sep=0.75pt]    {$x$};
% Text Node
\draw (127,37.4) node [anchor=north west][inner sep=0.75pt]    {$m$};

\end{tikzpicture}
\caption{We validate this controller approach against a PID controller that uses robust tuning.}  
\label{fig:spring-mass-damper}
\end{figure}

Throughout this work, we use the mass-spring-damper system as a baseline to evaluate the performance of the system identification model and control tasks. The mass-spring-damper system \citep{aastrom2007feedback} approximates the key motion and behavior of soft actuator systems \citep{dupont2009design, qiao2019dynamic, hainsworth2022simulating} as a linear system. 
Figure \ref{fig:spring-mass-damper} illustrates the mass-spring-damper system. 

Although it does not capture the nonlinear properties of soft actuators, showing that the neural-network based approach accurately captures the dynamics of such a system is a necessary condition for success at systems where a first-principled model is not available.

\section{Hardware testbeds}
\label{sec:hardware}

The physical soft actuator systems studied in this work are: 1) A tendon-based soft actuator with an internal network of optical sensors \citep{xu2019optical} and a Tensilica Xtensa LX6 (ESP32 Thing, Sparkfun)\citep{manzano2020high}, and 2) A folded HASEL-based, multilayered platform equipped with magnetic sensors and an ARM Cortex-M4f platform (Teensy 3.6, PJRC) \citep{sundaram2022embedded}.

The tendon-based platform in Figure \ref{fig:photographs}\textit{(A)} consists of a soft 3D-printed mesh that has a network of embedded \textit{LightLace} \citep{xu2019optical} stretchable sensors, organically weaved inside the soft mesh structure by hand during fabrication. The end-effector connects to two high torque servos (Dynamixel RX-64, Robotis) using Kevlar threads (Kevlar Fiber, Dupont) that serve as tendons. The motors pull the tendons, which in term moves the lattice's end effector in 3-dimensional coordinates. The undeformed mesh dimensions are 14 $\times$ 7 $\times$ 3 $cm^3$.  The task space of this end effector extends to 40.4 $\times$ 28.3 $\times$ 35.0 $mm^3$ and has the shape of a curved hull in three dimensions. The inputs to this system are the angular displacements of the servos from the manufacturer reference datum. The \textit{LightLace} sensor signal consists of eleven normalized sensor channels. Within the mesh, the \textit{LightLace} sensors may be subject to irreversible changes in position.

The HASEL-driven platform \citep{sundaram2022embedded} in Figure \ref{fig:photographs}\textit{(b)} has six folded-HASEL actuators and sensor units arranged in a triad configuration, with three actuators on the bottom layer and the remaining three on the top layer. The sensor units respond to changes in the magnetic properties of the actuating unit and we describe them in more detail in the section below. A 1.5-millimeter acrylic sheet covers the top sensors and carries the end-effector motion capture marker. The task space of the platform extends to a range of 23.0 $\times$ 20.2 $\times$ 19.2 $mm^3$ and has the shape of a point cloud around the end effector's marker in a neutral state. In this system, we have 18 sensor channels, three outputs for the end-effector positions, and six inputs to the HASEL actuators. 

\subsection{Sensing and Actuation.}
The tendon-based platform sensing method consists of a network of one-millimeter diameter, polyurethane fibers (Crystal Tec) \citep{xu2019optical} with input lines that carry light from infrared emitters (TSHA4400, Digi-key Electronics), with an intensity of 12mW/sr at 100mA, and output lines that carry coupled light to photodiodes (SFH 229, OSRAM Licht AG). The tubes are exposed to ambient light. The photodiodes capture changes of light intensity and produce an analog-to-digital converter (ADC) signal. An acrylic structure holds the soft mesh and the two actuators. One of the high torque servos has a vertical placement and the other servo motor has a horizontal placement to provide up-down, left-right, expressive, motion in the 3-dimensional task space. These high-torque servos receive commands from a circuit board (OpenCM9.04, Robotis) and receive power from a dedicated circuit board (OpenCM 485 Expansion board, Robotis). The power source is a commodity 12 V that connects to the wall.
 
Each sensing unit in the HASEL-driven platform consists of two components: a magnetic block and a 3-axis, off-the-shelf magnetometer (LIS3MDL, ST Electronics). The magnetic block is a mixture of platinum-catalyzed silicone (Ecoflex 00-30, Smooth On) and neo-powder (NQB-B+20441, Neo Magnequench), with a total mass of around 10 $g$ and dimensions of 50 $\times$ 50 $\times$ 5 $mm^3$. \cite{sundaram2022embedded} describes in more detail the process of manufacturing these sensors. We sandwich the HASEL unit between the magnetic block on top and the magnetometer at the bottom. The magnetometer senses the change of magnetic flux density as the magnetic block moves in conjunction with the HASEL's movement. The operating voltage range of the folded HASEL actuators is 0-8kV in this work, which we generate using a high voltage DC-DC converter (10A24-P30, Advanced Energy). To allow for independent control of each HASEL on the platform, we use six separate driving circuits to vary the applied voltage to each HASEL. The driving circuit includes a charging optocoupler (OZ100SG, Voltage Multipliers, Inc.) and a draining optocoupler to reduce the voltage. The changes in the height of six actuators fully describe the pose of the platform. 

Computationally, the tendon-based study in this work uses both the Python prototype and firmware code, yielding equal errors at the path following task. The HASEL-based studies in this work come strictly from firmware code and we study the mass-spring-damper analysis in simulation using our Python prototype. Profiling of this controller code in firmware in terms of memory, power and compute time is available in the \emph{Results} section.

\subsubsection{Software Testing and Unit Testing.} To validate firmware computation (C language), we use a combination of \emph{Valgrind} \citep{nethercote2007valgrind}, \emph{gdb} \citep{stallman1988debugging} and Simplified Wrapper and Interface Generator (\emph{SWIG}) \citep{beazley1996swig} to unit-test each of the firmware functions used in this work and also do the integration test. We unit-test neural network layers against Keras during \textit{nn4mc} development using SWIG. We set a relative tolerance of $10^{-5}$ to compare against the outputs of the equivalent layer in Keras for randomized layer properties during the development of each layer template. We also test the individual numerical gradients from the neural network models using the Tensorflow \emph{gradient tape} functionality for automatic differentiation for models with layers that have the functionality available and open to the public \citep{david2020tensorflow, tensorflowgradienttape}. We measure power and energy consumption of the control algorithm using a power monitor (Low-voltage Power Monitor, Monsoon Solutions, Inc.).

\subsubsection{Model training and Data Collection.} To collect training data from a four-camera motion capture system (Optitrack, Natural Point Inc) and train the neural network, we use a Linux personal computer with two graphic processing units (GPUs) that we then offload into firmware once it is trained.

\section{Problem Statement}
\label{sec:problem_statement}

This work considers a generalized, soft actuating platform that controls the position of an end-effector in $n$-dimensional space; thus, we define the state of the end-effector, or the output of the system, using $n$ variables, where $\mathbf{y} = \{y_0, ..., y_{n - 1}\}^T$ and is actuated by $m$ different actuators, such that the control input $\mathbf{u}$ is defined as $\mathbf{u} = \{u_0, ..., u_{m - 1} \}^T$. Embedded in this generalized system are $w$ channels of proprioceptive sensors, denoted as $\mathbf{l} = \{l_0, ..., l_{w - 1} \}^T$. The goal is to describe a desired geometric reference path, $\mathbf{y_{ref}} \in \mathbb{R}^{1 \times n}$ using information that is observable from the soft robot internal proprioception, i.e. no external observers.

We start from the nonlinear, differentiable function in Equation \ref{eq:model} that depends on a history of inputs to the system ($\tau \in \mathbb{R}^{1 \times n_d m}$), past predictions of outputs from the system ($\alpha \in \mathbb{R}^{1 \times d_d n}$), and instantaneous sensor signals ($\mathbf{l} \in \mathbb{R}^{1 \times w}$).

\begin{equation}
\label{eq:model}
\dot{\mathbf{y}}(t) = g(\mathbf{\tau}, \mathbf{\alpha}, \mathbf{l})
\end{equation}

Where $\mathbf{\tau}$ is a queue composed by $\{\mathbf{u}(t-n_d), ... ,\mathbf{u}(t)\}$;  $\mathbf{\alpha}$ is composed by $\{\mathbf{y}(t-d_d), ... ,\mathbf{y}(t-1)\}$ and $d_d$ and $n_d$, which are how far into the past the model looks at predictions and inputs respectively. We describe the necessary operations to update these queues in Figure \ref{fig:roll_deque} and the \emph{Operations} section. 

 $N$ is the prediction horizon, or the number of times the nonlinear discrete model will be recursively called in order to predict future outputs with $N_1$ and $N_2$, ($N_1 < N_2 \leq N$) being the start and end of the cost horizon defined in Equation \ref{eq:cost}. $N_c (\leq N )$ is the control horizon, which is used towards the prediction. To achieve the prediction matrix $\mathbf{\hat{Y}}\in \mathbb{R}^{N \times n}$, we modify the input vector $N$ times by recursively feeding a history of past predictions and also feeding a history of inputs $\mathbf{U} \in \mathbb{R}^{N_c \times m}$, because $N_c \leq N$ we copy the last row of $\mathbf{U}$ to predict past $N_c$ when $N_c < N$.

\section{Nonlinear Real-Time ML-based Controller}

The cost function $J$ in Equation \ref{eq:cost} is a combination between the standard, unconstrained, MPC cost function with additional terms that enforce smoothness in the optimal control inputs. $\mathbf{\Lambda} \in \mathbb{S}^{m\times m}$  and $\mathbf{Q} \in \mathbb{S}^{n \times n}$ are weighting matrices that penalize for the input changes and output square errors, respectively. 
\vspace{-3pt}
\begin{equation}
    \label{eq:cost}
    \begin{aligned}
        J & = \sum_{j=N_1}^{N_2} \| \mathbf{y}_{ref, j} - \hat{\mathbf{y}}_j \|_{\mathbf{Q}}^2   + \sum_{j=0}^{N_c} \| \Delta \mathbf{u}_j \|_{\mathbf{\Lambda}}^{2} \\ & + \sum_{i=1}^{m}  \sum_{j=1}^{N_c} \Bigg[ \frac{s}{u(n+j, i) + \frac{r}{2} - b} \\ & + \frac{s}{\frac{r}{2} + b - u(n+j, i)} - \frac{4}{r} \Bigg] 
    \end{aligned}
\end{equation}

The third term in Equation \ref{eq:cost} enforces smoothness in the control input, where $s$, $b$ and $r$ are tuning parameters. $s$ is usually a very low number \citep{soloway1996neural}.

\subsection{Newton-Raphson Solver.} In order to get an optimal control input, we frame the control problem as the solution to Equation \ref{eq:NR}, where $\frac{\partial^2 J}{\partial \mathbf{U}^2}(k)$ is the Hessian of the cost function at the current timestep $k$, and $\frac{\partial J}{\partial \mathbf{U}}(k)$ is the Jacobian of the cost function, where $\mathbf{U}$ as the vector composed by $\mathbf{U} = \{ \mathbf{u}_k, \mathbf{u}_{k+1}, ..., \mathbf{u}_{k+N_c}\}^T \in \mathbb{R}^{N_c \times m}$.

\begin{equation}
    \label{eq:NR}
    \begin{aligned}
    \frac{\partial^2 J}{\partial \mathbf{U}^2}(k)(\mathbf{U}(k+1) - \mathbf{U}(k)) = -  \frac{\partial J}{\partial \mathbf{U}}(k)
    \end{aligned}
\end{equation}

To solve for Equation \ref{eq:NR} we need two important components: 1) the first derivative of the cost function with respect to the input horizon and 2) the second derivative of the cost function with respect to the input horizon. We denominate these as the Jacobian and Hessian of the cost and describe them in Equations \ref{eq:cost1} and \ref{eq:cost2}.

\begin{equation}
    \label{eq:cost1}
    \begin{aligned}
        \frac{\partial J}{\partial \mathbf{U}}(k) & = -2 \sum_{j=N_1}^{N_2} (\mathbf{y}_{ref, j} - \hat{\mathbf{y}}_j)^T \mathbf{Q} \overbrace{\frac{\partial \hat{\mathbf{y}}_j}{\partial \mathbf{U}}}^{\text{Equation. \ref{eq:first_derivative_full}}}  \\ & + 2 \sum_{j=0}^{N_c} \Delta \mathbf{u}_j \mathbf{\Lambda} \circ \frac{\partial \Delta \mathbf{u}_j}{\partial \mathbf{U}} \\ & + \sum_{i=1}^{m}  \sum_{j=1}^{N_c}  \Bigg[ \frac{-s}{\Big[ u(n+j, i) + \frac{r}{2} - b  \Big]^2} \\ & + \frac{s}{\Big[ \frac{r}{2} + b - u(n+j, i) \Big]^2} \Bigg] \in \mathbb{R}^{N_c \times m}
    \end{aligned}
\end{equation}

Given the Kronecker delta operator $\delta(i, j) = \begin{cases} 0 & i \neq j \\ 1 & i = j\end{cases}$, the $h$'th row and the $j$'th column of $\frac{\partial \Delta \mathbf{u}_j}{\partial \mathbf{U}} \in \mathbb{R}^{N_c \times m}$ is computed by subtracting the Kronecker delta for two different dummy indices $\delta_{h, j} - \delta_{h, j - 1}$. Similarly, for the second derivative of the cost we use Equation \ref{eq:cost2}.

\begin{equation}
    \label{eq:cost2}
    \begin{aligned}
      \frac{\partial^2 J}{\partial \mathbf{U}^2}(k) & = \\
        & 2 \sum_{j=N_1}^{N_2}   \Big[ \mathbf{Q}  \Big(\frac{\partial \hat{\mathbf{y}}_j}{\partial \mathbf{U}} \circ \frac{\partial \hat{\mathbf{y}}_j}{\partial \mathbf{U}} \Big) - (\mathbf{y}_{ref, j} - \hat{\mathbf{y}}_j)^T \mathbf{Q} \frac{\partial ^2 \hat{\mathbf{y}}}{\partial \mathbf{U}^2} \Big]  \\ & + 2\sum_{j=0}^{N_c} \Big[ \mathbf{\Lambda}  \Big( \frac{\partial \Delta \mathbf{u}_j}{\partial \mathbf{U}} \circ \frac{\partial \Delta \mathbf{u}_j}{\partial \mathbf{U}} \Big)  + \Delta \mathbf{u}_j \mathbf{\Lambda} \circ \frac{\partial ^2 \Delta \mathbf{u}_j}{\partial \mathbf{U}^2} \Big] \\ & 
        + \sum_{i=1}^{m}  \sum_{j=1}^{N_c}  \Bigg[ \frac{2s}{\Big[ u(n+j, i) + \frac{r}{2} - b  \Big]^3} \\ & + \frac{2s}{\Big[ \frac{r}{2} + b - u(n+j, i) \Big]^3} \Bigg] \in \mathbb{R}^{N_c \times N_c}
    \end{aligned}
\end{equation}

\begin{figure*}
    \centering
    \input{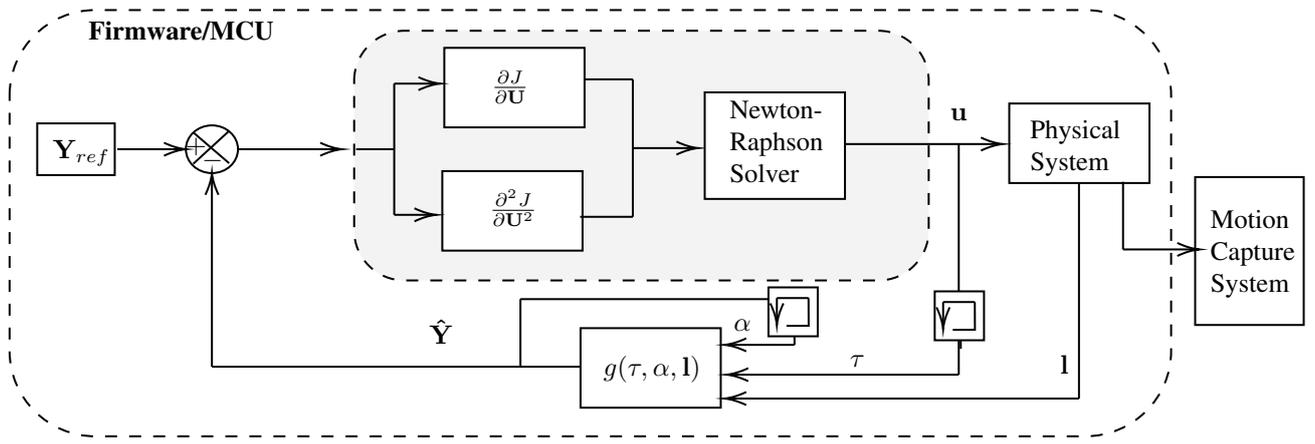} 
    \caption{A recursive, neural-based forward kinematic model, $g(\cdot)$, predicts outputs within a prediction horizon, $\mathbf{\hat{Y}}$. The algorithm computes a cost given a prescribed reference path. Then, it uses the Jacobian and Hessian of this cost structure towards an on-board Newton-Raphson solver, which computes the control input change to achieve the desired path. The motion capture system does not inform the controller of true states.}
    \label{fig:overview}
\end{figure*}

Where the factor $\frac{\partial ^2 \Delta \mathbf{u}_j}{\partial \mathbf{U}^2}$ always evaluates to zero. Equation \ref{eq:cost1} and \ref{eq:cost2} describe the expressions used toward the Jacobian and Hessian of the cost in Equation \ref{eq:cost}. Let $p = n_dm + d_dn+w$ be the length of the flattened input of the neural network and $\varepsilon$ be the differentiation stencil step length such that $[\mathbf{x}_{inputs} + \varepsilon\mathbf{I}] \in \mathbb{R}^{p \times p}$. Equation \ref{eq:big_x} is a matrix composed by repeating the latest neural network input vector of length $p$, $p$ times.  

\begin{equation}
    \label{eq:big_x}
    \begin{aligned}
   \mathbf{X}_j  & = \begin{bmatrix}\mathbf{x}_{inputs} \\ ... \\ \mathbf{x}_{inputs} \end{bmatrix} \in \mathbb{R}^{p \times n}
   \end{aligned}
\end{equation}

Equation \ref{eq:first_derivative1} is a temporary variable used to compute the gradient of the neural network model with respect to the neural network inputs using a second order finite differentiation stencil.

\begin{equation}
    \label{eq:first_derivative1}
    \begin{aligned}
   \Theta  & =  \frac{\partial \hat{\mathbf{y}}_j}{\partial \mathbf{x}_{inputs}}  \approx  \frac{g ( \mathbf{X}_j + \varepsilon\mathbf{I} ) - g ( \mathbf{X}_j - \varepsilon\mathbf{I})}{2\varepsilon} \\ & + \mathcal{O}(\varepsilon^2) \in  \mathbb{R}^{p \times n}
   \end{aligned}
\end{equation}

Once we have the gradients, we then compute the first derivative of the neural network with respect to the input matrix by taking the first $n_d \times m$ rows of the $\Theta$ matrix. In the firmware code implementation, the full gradient is never computed, but only the gradients necessary to get this matrix to avoid having to compute the rest of the rows and then discarding it. That is, $\Theta \in \mathbb{R}^{n_d  m \times n}$, which suffices to do the necessary compute and reduces the matrix sizes toward this computation. 

\begin{equation}
    \label{eq:reshape}
    \begin{aligned}
    \mathbf{O} = \Theta_{:n_d m, :}^T \in \mathbb{R}^{n \times n_d m} \rightarrow  \Theta_{:n_d m, :}^T \in \mathbb{R}^{ n \times n_d \times m}
    \end{aligned}
\end{equation}

Equation \ref{eq:reshape} symbolizes the reshape operation of the transpose of the first $n_d m$ rows of $\Theta$ from shape $n \times n_d m \rightarrow n \times n_d \times m$. 

\begin{equation}
    \label{eq:first_derivative_full}
    \begin{aligned}
    \frac{\partial \hat{\mathbf{y}}_j}{\partial \mathbf{U}} \approx \sum_{i = 0}^{n_d} \mathbf{O}(:, i, :) \in \mathbb{R}^{n \times m} 
    \end{aligned}
\end{equation}

For the second derivative ($\frac{\partial ^2 \hat{\mathbf{y}}}{\partial \mathbf{U}^2}$) we do a similar treatment to obtain this matrix. 

\begin{equation}
    \label{eq:second_derivative1}
    \begin{aligned}
   \chi  & =  \frac{\partial \hat{\mathbf{y}}_j^2}{\partial \mathbf{x}_{inputs}^2} \\ & \approx \frac{g ( \mathbf{X}_j + \varepsilon\mathbf{I} ) - 2g ( \mathbf{X}_j) + g (\mathbf{X}_j - \varepsilon\mathbf{I} )}{\varepsilon^2} \\ & + \mathcal{O}(\varepsilon^2) \in  \mathbb{R}^{p \times n}
   \end{aligned}
\end{equation}

\begin{equation}
    \label{eq:reshape2}
    \begin{aligned}
    \mathbf{O} = (\chi_{0:n_d m, :})^T \in \mathbb{R}^{n \times n_d m} \rightarrow  (\chi_{0:n_d m, :})^T \in \mathbb{R}^{ n \times n_d \times m}
    \end{aligned}
\end{equation}

\begin{equation}
    \label{eq:first_derivative_full2}
    \begin{aligned}
    \frac{\partial \hat{\mathbf{y}}_j^2}{\partial \mathbf{U}^2} \approx \sum_{i = 0}^{n_d} \mathbf{O}(:, i, :) \in \mathbb{R}^{n \times m} 
    \end{aligned}
\end{equation}

For speedup, we hold $\mathbf{X}_j = \mathbf{X}_{N-1} = \mathbf{X}$ in the firmware code, with $\mathbf{x}_{inputs}$ being the last recorded input to the neural network. This sacrifices fidelity with the recursive nature of $g(\cdot)$ as $N>1$ but allows for only one computation of the gradient of the recursive neural network. To get the inverse of the Hessian matrix from Equation \ref{eq:cost2} we use an on-board, lightweight, LU decomposition numerical module \citep{teukolsky1992numerical}. 

\subsection{Data Collection and Preprocessing}
\label{sec:collection}

Regardless of the system's actual implementation, this method requires a starting dataset that contains sample inputs, and corresponding sensor signals and ground-truth position measurements. 

\subsubsection{Mass-spring-damper.} We define the output state variable as $\mathbf{y} = [x_0]$, the control input is an external force divided by mass $\mathbf{u} = [\frac{F_{external}}{m}]$ and $w=0$ because there are no sensor signals fed back into the controller. We set model parameters as $k=$ 40 $\frac{N}{m}$, $c =$ 0.5 $\frac{N\cdot s}{m}$, $m =$ 0.1 $kg$. Equation \ref{eq:spring-mass-damper} is the equation of motion for this system in state space, where $x_0 = x, x_1 = \dot{x}$. We generate synthetic data by inputting a specific force of the form $u(t) = 1000 sin(t) cos(t)$ in $\frac{N}{kg}$ for times 0 to 2000 with increments of 1 millisecond. Then, we use \emph{odeint} \citep{ahnert2011odeint} to solve for the forward kinematic of the end-effector of the mass-spring-damper system.

\begin{equation}
    \label{eq:spring-mass-damper}
   \mathbf{\dot{x}} = \begin{bmatrix} x_1 \\ -\frac{k}{m}x_0 -\frac{c}{m}x_1 + \frac{F_{external}}{m} \end{bmatrix} 
\end{equation}

Figure \ref{fig:forward_kino}(a) displays the learned forward kinematic model applied to the mass-spring-damper system. We compare our control approach with a linear controller approach in Figure \ref{fig:pid_vs_nlsoft}.

\subsubsection{Tendon- and HASEL-driven platforms.}
To collect the training dataset, we record time series data from the sensor signals, the motion capture system, and the servo inputs $\mathbf{u}$. We make the tendon-driven platform follow a sequence of prescribed inputs (up-down, top-bottom and mixed) inputs. This sequences is repeated 50 times. We then prepare the data depending on the values of $n_d$ and $d_d$ using Equation \ref{eq:data}. This results in a data size of 3.8 million samples, which are then reorganized depending on $n_d$ and $d_d$. The signals have a discretization of an average step duration of $8.\bar{3}$  $ms$ or 120 $Hz$. For the HASEL-driven platform, we perform a similar data collection procedure as the tendon-driven platform, the data collection rate is 200 $Hz$. For training, this data is down-sampled to 130 $Hz$ to account for the solution time of the algorithm. 

We then prepare the datasets for training by shifting the matrix of actuator inputs $n_d$ times, shifting the matrix of forward kinematic outputs $d_d$ times, and appending instantaneous sensor signals. Throughout this work and across all platforms, we set $n_d = d_d = 2$. Equation \ref{eq:data} represents the steps toward successfully making these data copies, where $q$ is the number of rows collected from the data collection step, $\mathbf{\mho} \in \mathbb{R}^{q \times m}$ in this context is the history of all control inputs sent to the system, $\mathbf{\Upsilon} \in \mathbb{R}^{q \times n}$, and the recorded signals from the sensors $\mathbf{S} \in \mathbb{R}^{q \times w}$. The subscripts in Equations \ref{eq:data} are ranges of rows from the original matrices; all the columns from $\mathbf{\Upsilon}$, $\mathbf{\mho}$ and $\mathbf{S}$ are appended onto $\mathbf{X} \in \mathbb{R}^{(q - max(n_d, d_d)) \times p}$, where $p = n_dm + d_dn+w$.

\begin{equation}
    \label{eq:data}
    \begin{aligned}
    \mathbf{X}  = & [\mathbf{\mho}_{0:q - n_d}, \mathbf{\mho}_{1:q - n_d - 1}, ..., \\ & \mathbf{\mho}_{n_d-1:q}, \mathbf{\Upsilon}_{0:q - d_d}, \mathbf{\Upsilon}_{1:q - d_d -1}, ..., \\  & \mathbf{\Upsilon}_{d_d:q}, \mathbf{S}_{n_d:q}]
    \end{aligned}
\end{equation}

We use $\mathbf{y}_{true} = \mathbf{\Upsilon}_{n_d:q}$ for true labels. During experiment time, we append the data observable from the soft robot using the operations described in the section \emph{Operations}.

\subsection{Learning Forward Kinematics}

In \cite{manzano2020high} we compare three core model structures for the forward kinematics of the soft actuator. The Gated Recurrent Unit (GRU)-based neural network model outperforms its counterparts (LSTM, RNN) in terms of accuracy and memory usage. In this work, we train a structure consisting of an input layer of size $p$, one GRU layer with 5 units, one fully connected (FC), rectified linear unit (ReLU) layer with 5 outputs, and another FC layer with a hyperbolic tangent activation of $n$ outputs. For speedup, the model we deploy for the HASEL-based platform in hardware is a fully connected neural network with $p$ inputs, a hidden ReLu layer of 5 nodes, another ReLU layer of 5 nodes and the output is a hyperbolic tangent layer of 3 nodes. 

 We train the neural network regression model to predict the output transition between $k-1$ and $k$ by partitioning the datasets into ten-fold time series splits \citep{medar2017impact}. This allows for the GRU to successfully capture in its memory time series data similar to what it will capture in real time during experiment time. As testing set, we use the last time series split, which represents the last bucket of time series test set data after performing the splits. Data is normalized and shifted in order to make all the motions in task space to surround the neutral point of the end-effector, which we use as the initial guess of the Newton-Raphson method. Inputs are min-max normalized to be bounded in the range $\{-0.5;0.5\}$ and sensor signals are also min-max normalized to be in the range $\{-0.5;0.5\}$.

\subsubsection{Mass-spring-damper.} The test set predictions for the forward kinematics of the spring-mass-damper system using a treatment similar to the above platforms, we achieve a test set mean-squared error of $5.5\times10^{-4} m$ on the last time series partition. Results are displayed in Figure \ref{fig:forward_kino}\textit{(a)}.

\subsubsection{Tendon-based Actuator.} A sample of the forward kinematics test set results are displayed in Figure \ref{fig:forward_kino}\textit{(b)}, where the first row indicates the motor inputs $u_0$ and $u_1$ in radians, $y_0, y_1$, and $y_2$ are the output states in millimeters. The average test set error mean squared error is 0.12 $mm$.

\subsubsection{HASEL-based Actuator.} 
\begin{figure}
    \hfill
    \centering
    \begin{subfigure}[t]{\columnwidth}
        \centering
        \begin{tikzpicture}
        \begin{scope}
        \node (image) at (0,0) {\includegraphics[width=1
        \columnwidth]{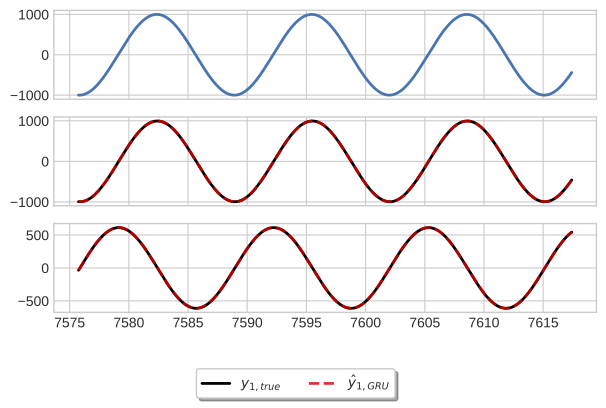}};
        \end{scope}
        \node (note) at (0.0,3.3) {\textbf{Mass-Spring-Damper Forward Kinematic}};
        \node (note) at (0.,2.9) {\textbf{Predictions on Test Set}};
        \node[rotate=90] (note) at (-4.3,0.2) {$x_0$ $[m]$};
        \node[rotate=0]  (note) at (-4.2,3.3) {\textbf{(a)}};
        \node[rotate=90] (note) at (-4.3,-1.3) {$x_1$ $[m]$};
        \node[rotate=90] (note) at (-4.3,1.8) {$u$ $[\frac{N}{kg}]$};
        \node[rotate=0]  (note) at (0.0,-2.) {Time [s]};
        \end{tikzpicture}
    \end{subfigure}
    \hfill
   \begin{subfigure}[t]{\columnwidth}
        \centering
        \begin{tikzpicture}
        \begin{scope}
        \node (image) at (0,0) {\includegraphics[width=\columnwidth]{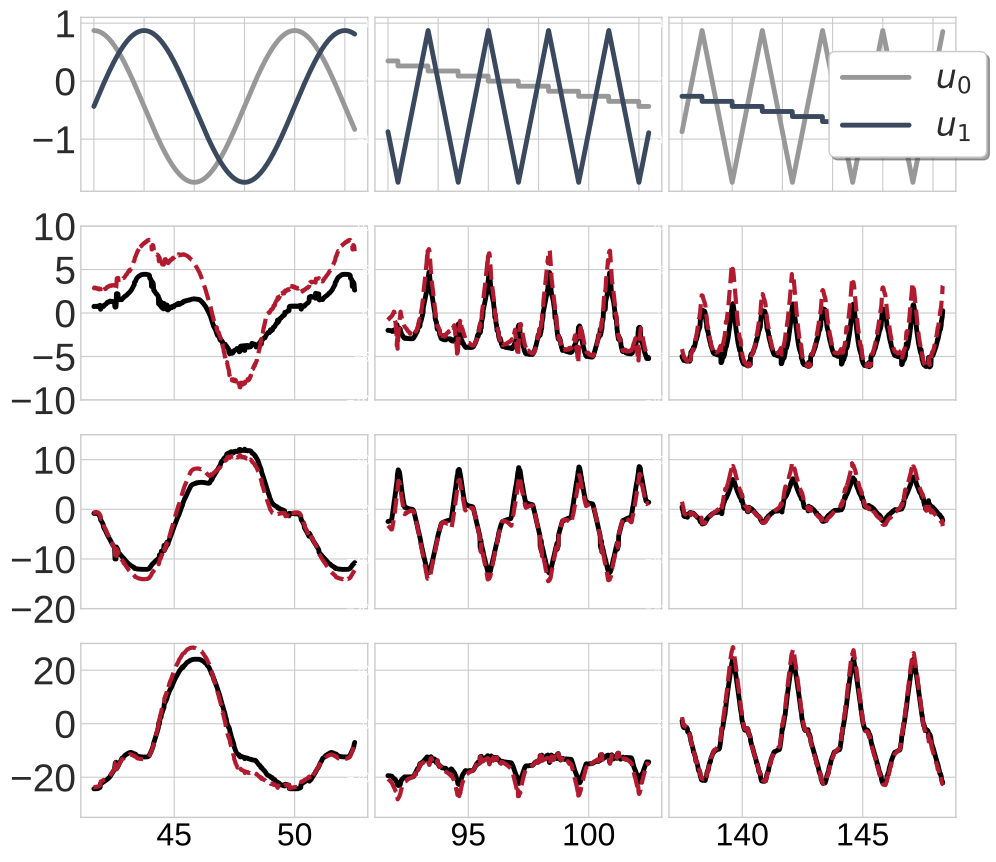}};
        \end{scope}
        \node (note) at (-0.3,4.2) {\textbf{Tendon-based Forward Kinematic}};
        \node (note) at (-0.3,3.8) {\textbf{Predictions on Test Set}};
        \node[rotate=90] (note) at (-4.4,1.3) {$y_0$ $[mm]$};
        \node[rotate=0] (note) at (-4.2,4.2) {\textbf{(b)}};
        \node[rotate=90] (note) at (-4.4,-0.3) {$y_1$ $[mm]$};
        \node[rotate=90] (note) at (-4.4,-2.5) {$y_2$ $[mm]$};
        \node[rotate=90] (note) at (-4.4,3.) {$u$ $[rad]$};
        \node[rotate=0] (note) at (0.,-3.9) {Time [s]};
        \end{tikzpicture}
    \end{subfigure}
   \hfill
   \centering
    \begin{subfigure}[t]{\columnwidth}
        \centering
        \begin{tikzpicture}
        \begin{scope}
        \node (image) at (0,0) {\includegraphics[width=1
        \columnwidth]{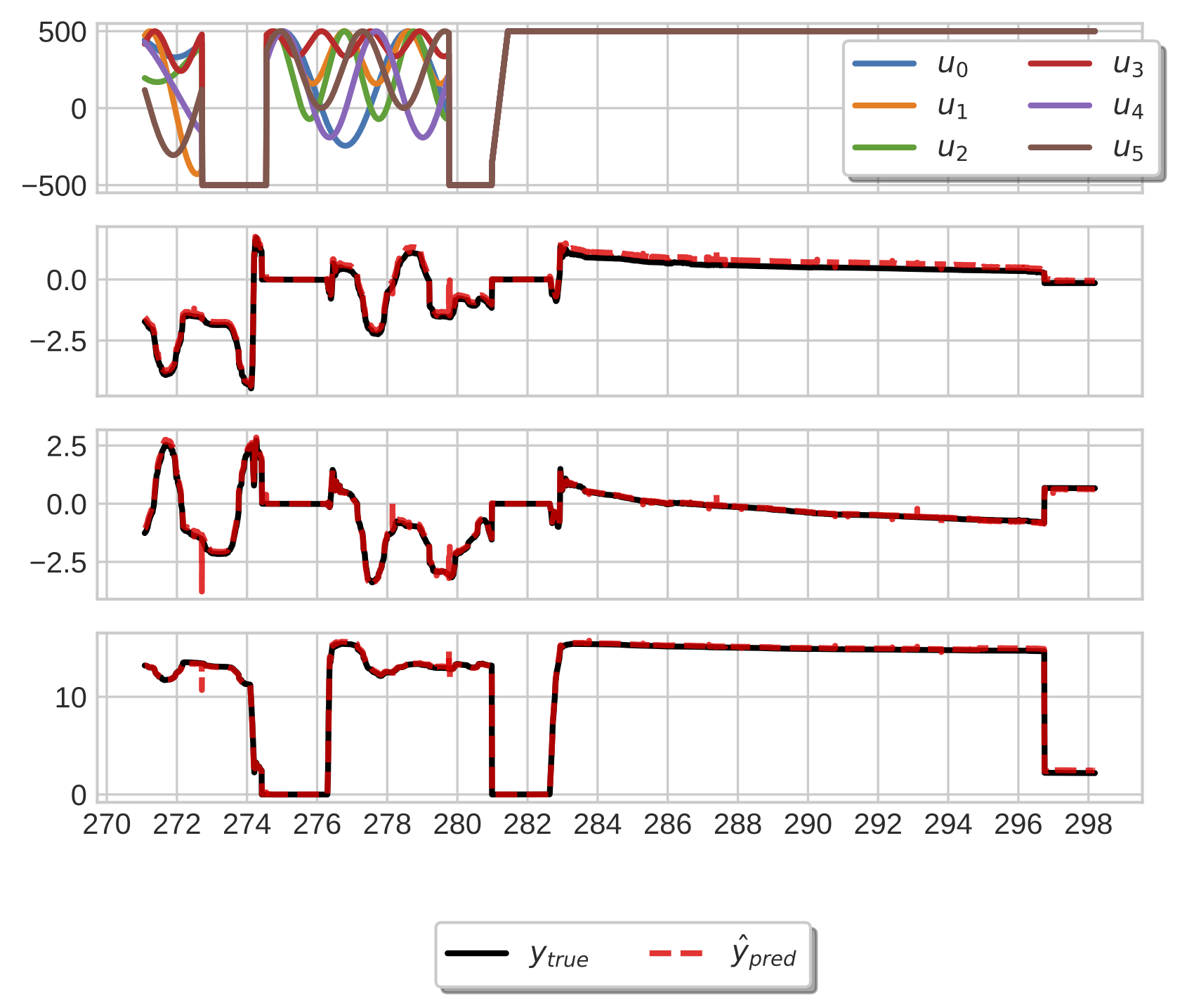}};
        \end{scope}
        \node (note) at (0.1,4.2) {\textbf{HASEL-based Forward Kinematic}};
       \node (note) at (0.1,3.8) {\textbf{Predictions on Test Set}};
        \node[rotate=90] (note) at (-4.2,1.4) {$y_0$ $[mm]$};
        \node[rotate=0] (note) at (-4.2,4.2) {\textbf{(c)}};
        \node[rotate=90] (note) at (-4.2,0.) {$y_1$ $[mm]$};
        \node[rotate=90] (note) at (-4.2,-1.6) {$y_2$ $[mm]$};
        \node[rotate=90] (note) at (-4.2, 2.9) {$u$ $[pwm]$};
        \node[rotate=0] (note) at (0.0,-2.7) {Time [s]};
        \end{tikzpicture}
    \end{subfigure}
    \caption{Forward kinematic models on testing set prediction compared to ground truth for the three systems studied. Starting from data on the system input, ground-truth output and signals, a neural network learns to predict the transition between discrete time step $k$ and $k + 1$. Ten-fold time series splits allow to validate the learning of the forward kinematics and the last time series split testing set prediction is illustrated in this figure.}
    \label{fig:forward_kino}
\end{figure}
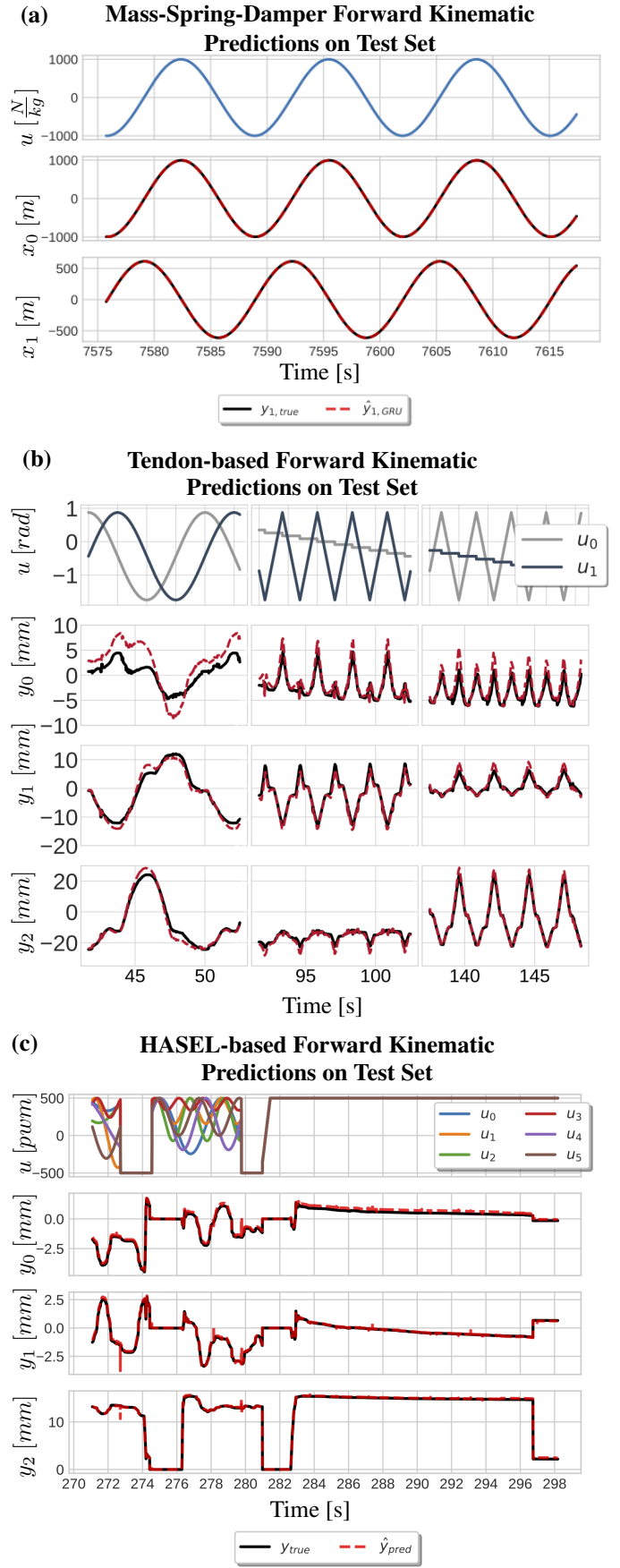

The forward kinematic predictions for a sample of the the test set from the HASEL-driven platform are displayed in Figure \ref{fig:forward_kino}\textit{(c)}. True data is a subset of the batch collected in STEP 1 of this approach. The offline test set mean squared error on the last time series split is $7.3 \times 10^{-2}$ $mm$.

\subsection{Compiling and deploying the controller}
\label{sec:deploy}

The controller described here is included with the open-source software. Deploying this controller with a trained model involves two steps: Neural network compilation, tuning and integration. For the neural network compilation we use code that \emph{nn4mc} generates and integrate the neural network code files into the rest of the firmware code. The tuning is done at the platform itself and consists of adjusting $\mathbf{Q}$, $\mathbf{\Lambda}$, $s, b, r$ to produce the desired behavior for path tracking. We achieve this tuning by monitoring the results of the controller in the MCU using serial output and measuring the difference between the predicted and the true trajectories as illustrated in Figure \ref{fig:predicted_hasel}. 

\subsection{Onboard Operations and Data Structures}
\label{sec:operations}

To make this controller lightweight, the firmware executes the following operations using the following data structures.

\subsubsection{Matrix2} \texttt{struct Matrix2} is the internal representation of matrices within the microcontroller code. This representation allows for allocation, linear algebra operations. \texttt{Matrix2} stores all 2-dimensional matrices as flat arrays into the microcontroller, these are later dynamically allocated and freed depending on the variable's utilization.

\begin{figure}[H]
\begin{tikzpicture}[font=\ttfamily,
array/.style={matrix of nodes,nodes={draw, minimum size=7mm, fill=yellow!30},column sep=-\pgflinewidth, row sep=0.5mm, nodes in empty cells,
row 1/.style={nodes={draw=none, fill=none, minimum size=3mm}},
row 1 column 1/.style={nodes={draw}},
row 1 column 2/.style={nodes={draw}},
row 1 column 3/.style={nodes={draw}}}]

\matrix[array] (array) {
0 & 1 & 2 & 3 & 4 & 5 & $...$ & -bsize  & $...$ & -1 \\
  &   &   &   &   &   &   &   &   &  \\};
\node[draw, fill=gray, minimum size=5mm] at (array-2-9) (box2) {};
\node[draw, fill=gray, minimum size=5mm] at (array-2-10) (box3) {};
\node[draw, fill=gray, minimum size=5mm] at (array-2-8) (box4) {};

\begin{scope}[on background layer]
\fill[yellow!10] (array-1-1.north west) rectangle (array-1-10.south east);
\end{scope}

\draw[->]([yshift=-3mm]array-2-1.south west) -- node[below] {fixed queue length (nd * m)} ([yshift=-3mm]array-2-7.south east);

\draw (array-1-1.north)--++(90:3mm) node [above] (first) {};
\draw (array-1-2.north)--++(90:3mm) node [above] (first) {new elements};
\draw (array-1-3.north)--++(90:3mm) node [above] (first) {};
\node [align=center, anchor=south] at (array-2-9.north west|-first.south) (8) {elements \\ to be popped};
\draw (8)--(box2);
\draw (8)--(box3);
\draw (8)--(box4);

\end{tikzpicture}
\vspace{-10pt}
\caption{Roll operation on a queue. This is the equivalent to a simultaneous \texttt{push()} and a \texttt{pop()}.} 
\label{fig:roll_deque}
\end{figure}
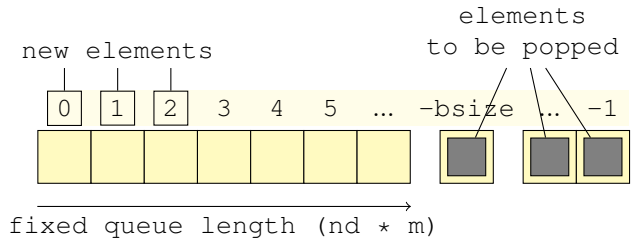

\subsubsection{Rolling} \texttt{roll(queue, start, end, bsize)}. 
The in-place rolling operation allows for the input vector to properly offset old data in order to give space to new incoming data. Rolling is equivalent to pushing and popping \texttt{bsize} elements from a queue. The rolling operation in the firmware code requires specified bounds (\texttt{start} and \texttt{end}) within the input vector because we only roll the elements within those bounds. Figure \ref{fig:roll_deque} exemplifies rolling three elements from a queue with bounds between \texttt{start = 0} and \texttt{end = nd * m}. All queues in this work are used as arrays where the first \texttt{bsize} elements are the most recent and the last \texttt{bsize} elements are the least recent.

\subsubsection{Input Vector Computation.} When the controller applies an optimal control input vector $\mathbf{u}_{t}$ of size $m$ and our past neural network prediction $\mathbf{y}_{t - 1}$, we also collect instantaneous sensor signals $\mathbf{l_{t}}$. Building the input vector is a successive chain of rolling operations like the pseudo-code below in Algorithm \ref{alg:input_vector}.

\begin{algorithm}
\caption{Input Vector Building (In-place)}\label{alg:input_vector}
\begin{algorithmic}
\Require $\mathbf{x}_{input} \in \mathbb{R}^{1 \times p}$
\State \texttt{roll}$(\mathbf{x}_{input}, 0, n_d * m - 1, m)$
\State $\mathbf{x}_{input}[0:m] \gets \mathbf{u}_t$
\State \texttt{roll}$(\mathbf{x}_{input}, nd*m, dd * n + nd*m - 1, n)$
\State $\mathbf{x}_{input}[nd*m:nd*m + n] \gets \mathbf{y}_{t - 1}$ 
\State $\mathbf{x}_{input}[-w:-1] \gets \mathbf{l}_t$ 
\end{algorithmic}
\end{algorithm}

\vspace{-10pt}

\section{Results}
\label{sec:results}

\begin{figure}
 \begin{tikzpicture}
        \begin{scope}
        \node (image) at (0,0) {\includegraphics[width=0.95\columnwidth]{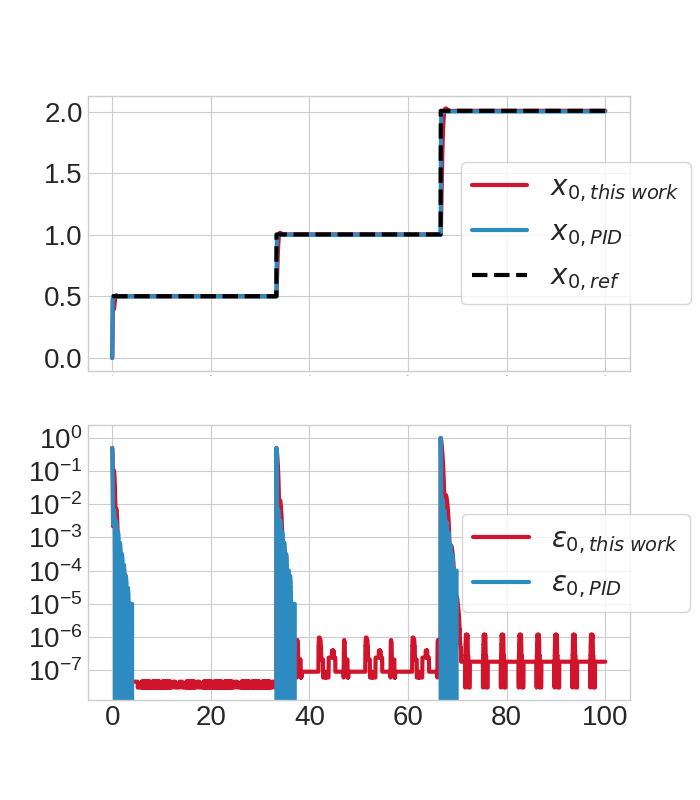}}; 
        \end{scope}
        \node[rotate=90] (note) at (-4.3,2) {$x_0$ [m]};
        \node[rotate=0] (note) at (-4.3, 4) {\textbf{(a)}};
        \node[rotate=0] (note) at (-4.3, 0) {\textbf{(b)}};
        \node[rotate=90] (note) at (-4.3, -1.9){$\varepsilon_0$ [m]};
        \node[rotate=0] (note) at (0, 4) {\textbf{Mass-Spring-Damper Simulation}};
        \node[rotate=0] (note) at (0,-4.3) {Percent Completion [\%]};
\end{tikzpicture}

\caption{We compare the tracking of a 1-dimensional reference position of the mass-spring-damper between our approach (simulation) and a PID approach. Part (a) is the tracking of the reference position; part (b) is the tracking error in logarithmic scale, measured in absolute difference between reference and true position in simulation. The fraction completed is a percentage along the task completion $\Big[ \frac{t}{t_{final}}*100 \Big]$.} 
\label{fig:pid_vs_nlsoft}
\end{figure}
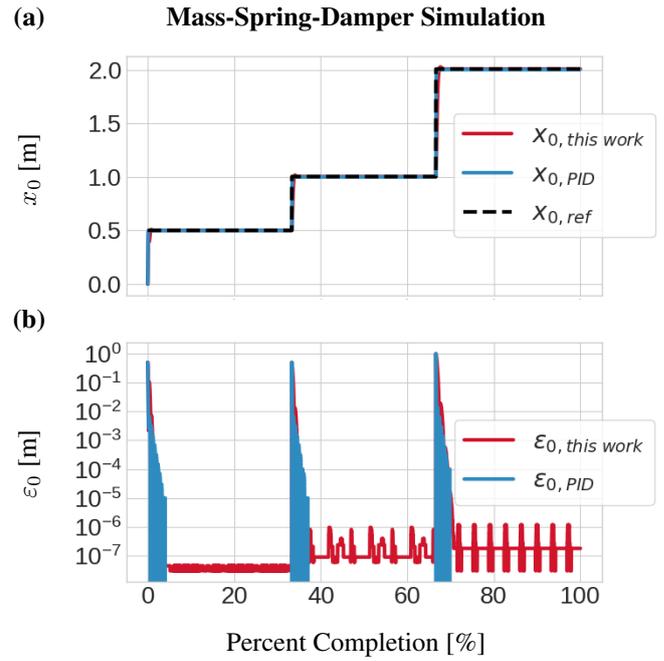

Figure \ref{fig:pid_vs_nlsoft} displays the results of the approach from this work in simulation compared to a PID controller that uses robust tuning. The nominal tunings for the PID controller are $K_p = 1.93$, $K_i = 4.01$, $K_d = 5.99$. The nonlinear control tunings for this system are $N = 1$, $Q = \begin{bmatrix} 1 \end{bmatrix}$, $C = diag\{1, 0\}$, $\Lambda = \begin{bmatrix} 1 \end{bmatrix}$, $s = 10^{-20}, b = 10^{-5}, r = 4\times 10^2$. Other values that are not tuning values are $\mathbf{x}_{k = 0} = [0., 0.]^T$, $\epsilon = 1\times 10^{-3}$, $n_d = d_d = 2$. In this simulation, we use a larger scale reference path (steps of 0.5m, 1m and 2m). For our nonlinear control approach, the rise time from 0\% to 90\% is 0.11s, 0.33s and 0.43s; the percentage overshoot is 1.8\%, 3.2\% and 2.5\% at each step; and the steady-state error is 1.4mm, 2.8mm and 5.7mm, the total time elapsed ($t_{final}$) for the task is 80s. For the PID automated approach, the rise time between 0-90\% is 0.94s for the three steps; the percentage overshoot is 0.64\%, 0.64\% and 0.13\%; the steady-state error is 0mm in this simulation, $t_{final}$ is 1000s.

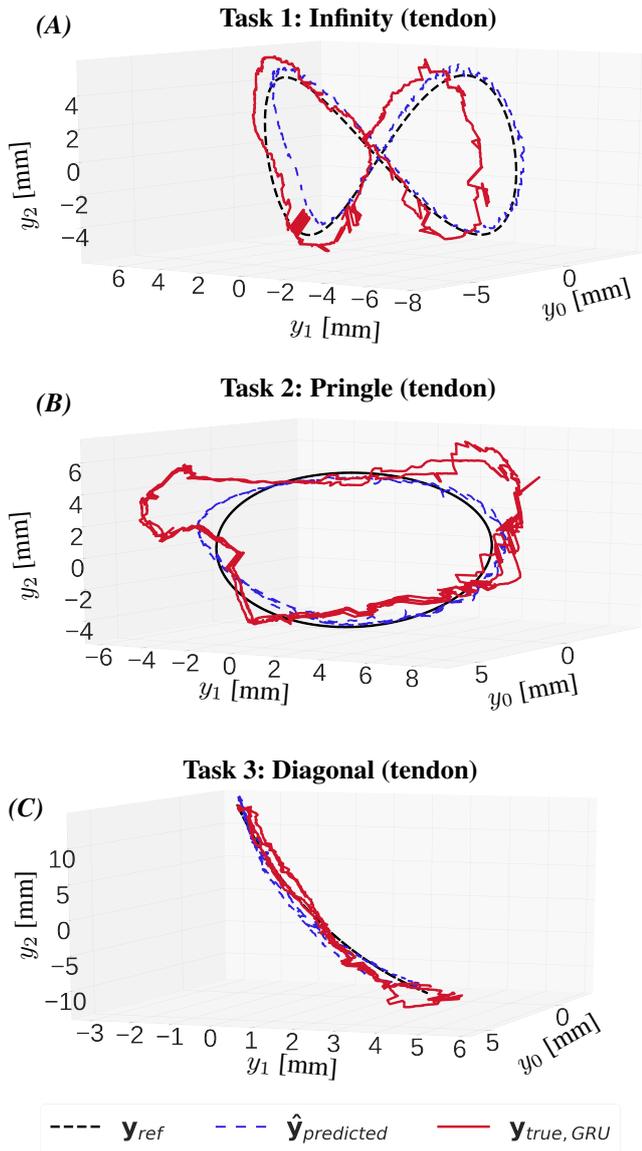
\begin{figure}
    \centering
    % true Eight
    \begin{subfigure}[t]{\columnwidth}
        \centering
        \begin{tikzpicture}
        \begin{scope}
        \node (image) at (0,0) {\includegraphics[width=\columnwidth, trim={6cm 1cm 6cm 1cm},clip]{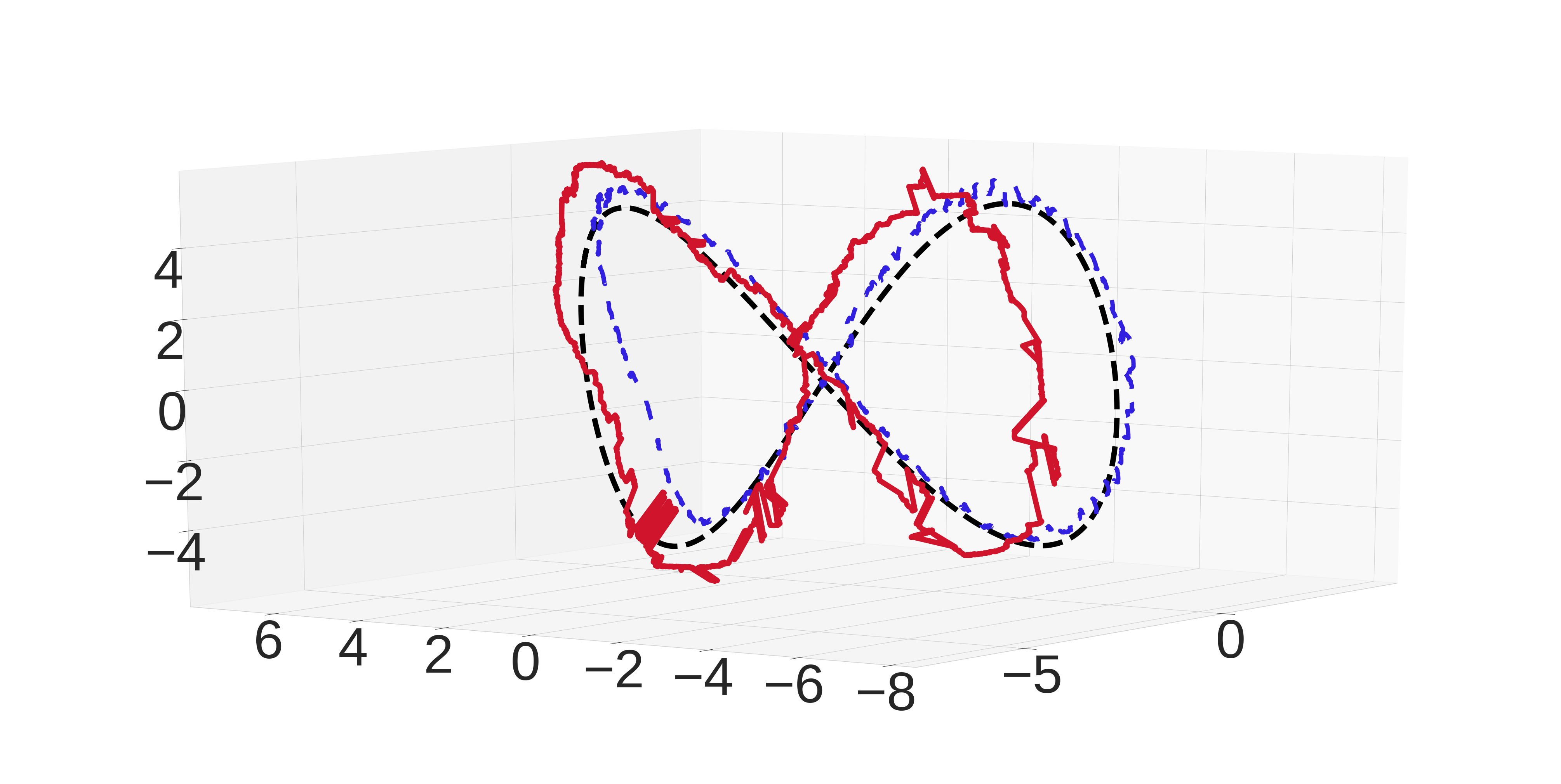}}; 
        \end{scope}
       \node[rotate=95] (note) at (-4.4,-0.3) {$y_2$ [mm]};
       \node[rotate=-5] (note) at (-0.3,-2.2) {$y_1$ [mm]};
       \node[rotate=15] (note) at (3,-1.8) {$y_0$ [mm]};
       \node[rotate=0] (note) at (0,1.9) {\textbf{Task 1: Infinity (tendon)}};
       %  \node[rotate=10] (note) at (2.2,-2) {$y_0$ [mm]};
          \node (note) at (-4,1.8) {\textit{\textbf{(A)}}};
        \end{tikzpicture}
    \end{subfigure}
    \centering
    % true pringle
    \begin{subfigure}[t]{\columnwidth}
        \centering
        \begin{tikzpicture}
        \begin{scope}
        \node (image) at (0,0) {\includegraphics[width=\columnwidth, trim={6cm 1cm 6cm 1cm},clip]{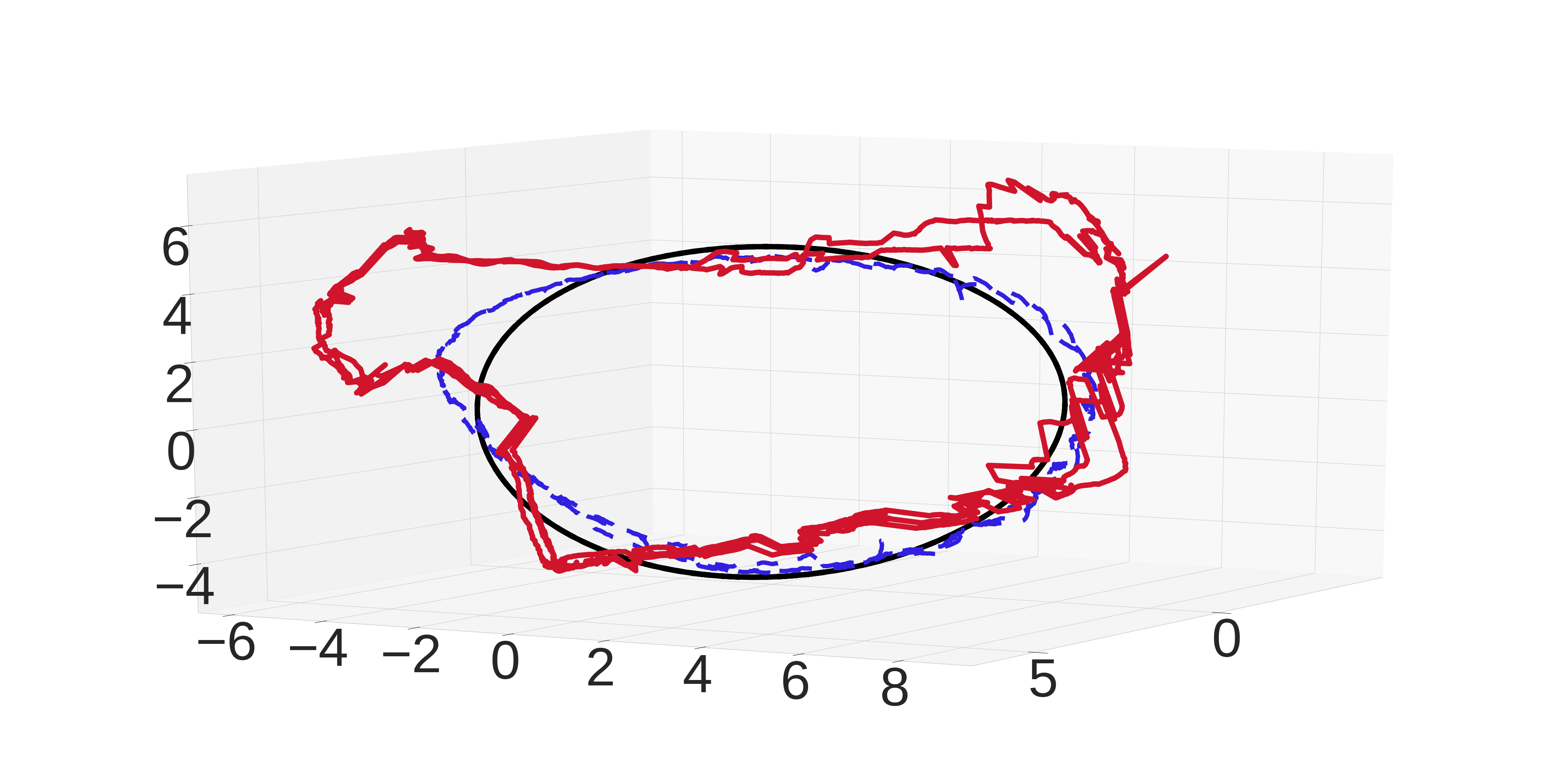}}; 
        \end{scope}
        \node[rotate=95] (note) at (-4.4,-0.2) {$y_2$ [mm]};
        \node[rotate=0] (note) at (0,2.0) {\textbf{Task 2: Pringle (tendon)}};
        \node[rotate=-5] (note) at (-1.5,-2.) {$y_1$ [mm]};
        \node[rotate=10] (note) at (2.3,-2.0) {$y_0$ [mm]};
        \node (note) at (-4,1.8) {\textit{\textbf{(B)}}};
        \end{tikzpicture}
    \end{subfigure}
    \centering
    % true diagonal
    \begin{subfigure}[t]{\columnwidth}
        \centering
        \begin{tikzpicture}
        \begin{scope}
        \node (image) at (0,0) {\includegraphics[width=\columnwidth, trim={6cm 0cm 6cm 1cm},clip]{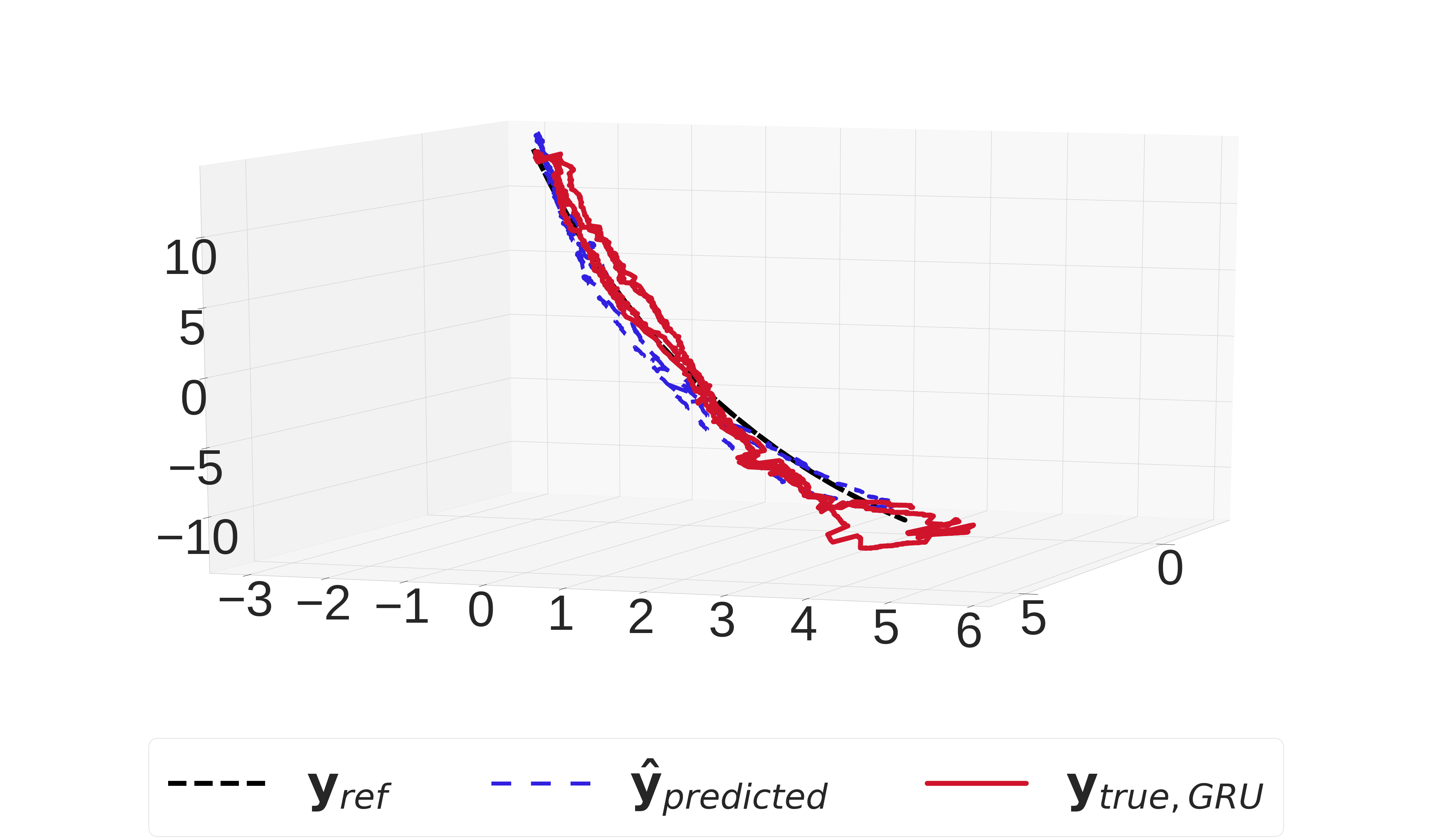}}; 
        \end{scope}
        \node[rotate=90] (note) at (-4,0.4) {$y_2$ [mm]};
        \node[rotate=-5] (note) at (-0.5,-1.6) {$y_1$ [mm]};
        \node[rotate=0] (note) at (0,2.3) {\textbf{Task 3: Diagonal (tendon)}};
        \node[rotate=30] (note) at (3.0,-1.3) {$y_0$ [mm]};
        \node (note) at (-4,1.8) {\textit{\textbf{(C)}}};
        \end{tikzpicture}
    \end{subfigure}
    \caption{Three-dimensional path tracking of the tendon-driven platform for three different paths as measured using the motion capture system in $\mathbf{y}_{true,GRU}$. We define the task space as $\mathbf{y} = \{y_0, y_1, y_2\}$ and the path predictions as $\mathbf{\hat{y}}_{predicted}$.}
    \label{fig:path_tracking}
\end{figure}

We measure the true path followed by the end-effector using the motion capture system for the three paths followed: \emph{Infinity} in Figure \ref{fig:path_tracking}(a); \emph{Pringle} in Figure \ref{fig:path_tracking}(b); \emph{Diagonal} in Figure \ref{fig:path_tracking}(c) for the tendon-based platform and we monitor the path tracking from serial communication at the firmware of the HASEL-based platform to generate Figure \ref{fig:path_tracking}.

\subsection{Path Following}

\subsubsection{Tendon-driven path following.} We test our controller on the path following tasks that we named: 1) \emph{Infinity}, 2) \emph{Pringle} and 3) \emph{Line}, and which are defined below. We describe the \textit{Infinity} reference trajectory as $\mathbf{y_{ref}}(t) = 
\{A sin^2(\omega t ) + y_{0, 0}, 
 B sin(\omega t ) cos(\omega t ) + y_{1, 0},  
 C sin(\omega t ) + y_{2, 0} \}^T$, where $\omega$ 
 is the frequency of the wave described by this path. Parameters $A$ and $B$ are unitless multipliers that indicate amplitude and extension of the reference path geometry. Figure \ref{fig:path_tracking} shows the difference between the true location of the end effector compared with the estimated location of the end effector for three different recurrent models used in this work. \textit{Pringle} is defined as a hyperbolic paraboloid $\mathbf{y_{ref}}(t) = \{
A (y_2^2 / B^2 - y_1^2 / C^2) + y_{0, 0},
Bcos(2\pi  \omega) t + y_{1, 0}, 
Csin(2\pi  \omega) t + y_{2, 0}\}^T$, 
and \textit{Line} is defined as $\mathbf{y_{ref}}(t) = 
\{A (y_2^2 / B^2 - y_1^2 / C^2) + y_{0, 0}, 
Bsin(2\pi  \omega t+ 10^{-6} t^2) + y_{1, 0}, 
Csin(2\pi  \omega t+ 10^{-6} t^2) + y_{2, 0} \}^T$. We achieve a mean root-mean-squared (RMS) error of $1.28 \pm 0.23$ mm for the \emph{Infinity} path, $1.56 \pm 0.58$ mm for the \emph{Pringle} path , and $2.02 \pm 1.19$ mm at the \emph{Diagonal} path. Variable $\omega$ is a wavelength variable that controls how fast the reference trajectory changes with respect to the discretized time step $k$. 

The tuning parameters used for the tendon-driven platform were: $\mathbf{Q} = diag([1 \times 10^{-3}, 5\times 10^3, 2\times 10^3])$, $\mathbf{\Lambda} = \mathbf{I}_{2 \times 2}$, $\epsilon = 1./120.$, $N = 3$, $N_c = 3$, $s = 1\times 10^{-20}$, $b = 1 \times 10^{-10}$, $r = 4 \times 10^5$, $n_d = d_d = 2$.

\begin{figure}
    \centering
     \begin{tikzpicture}
       \begin{scope}
       \node (image) at (0,0) {\includegraphics[width=\columnwidth, trim={6cm 0cm 6cm 1cm},clip]{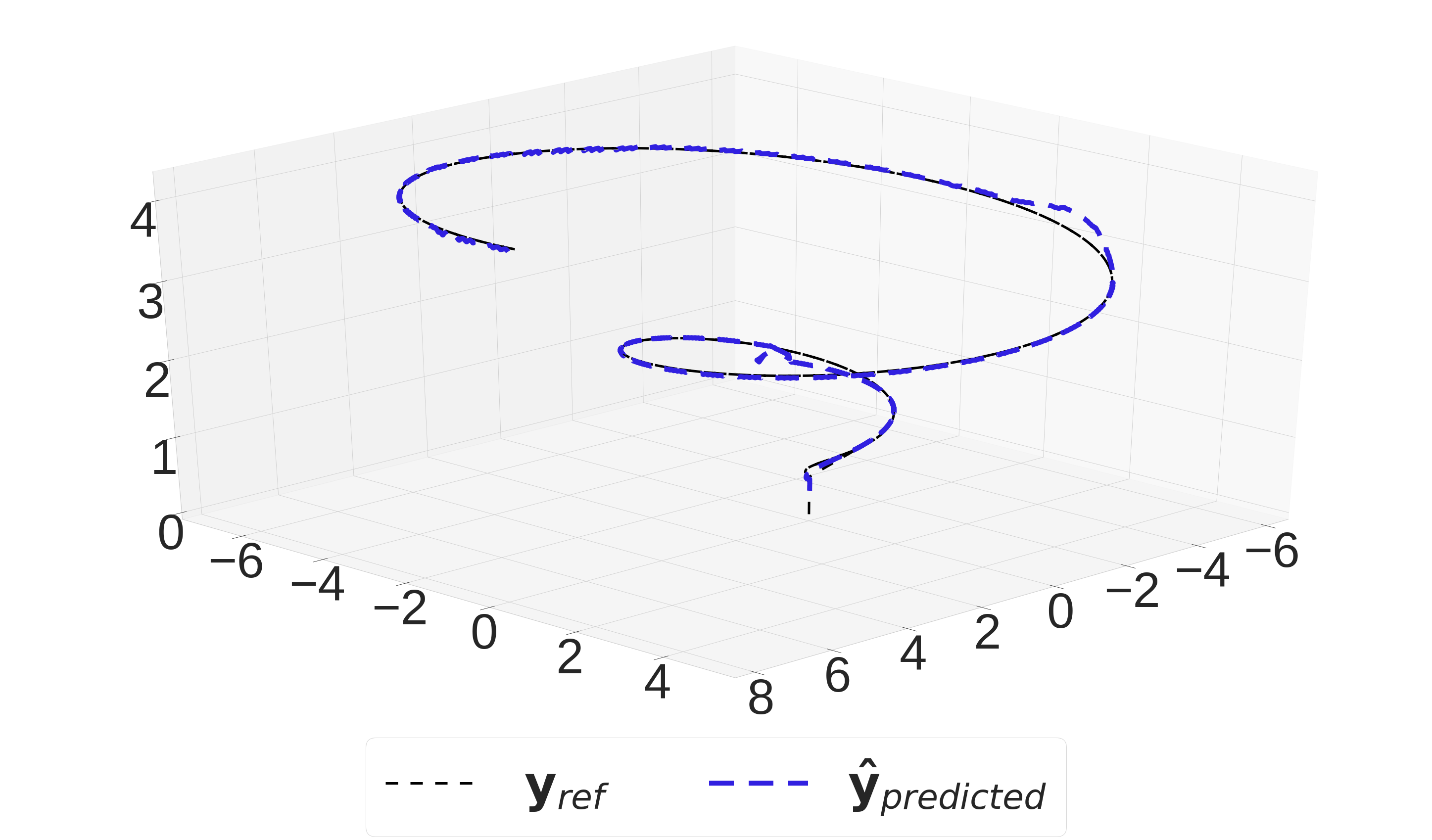}}; 
       \end{scope}
       \node[rotate=95] (note) at (-4.5,0.2) {$y_2$ [mm]};
       \node[rotate=-15] (note) at (-2.0,-1.8) {$y_0$ [mm]};
      \node[rotate=0] (note) at (0,2.5) {\textbf{Task 4: Swirl (HASEL)}};
       \node[rotate=15] (note) at (2.2,-1.8) {$y_1$ [mm]};
    \end{tikzpicture}
    \caption{Recorded output predictions from the computational platform (ARM Cortex-M4f). The neural network predicts the forward kinematics for the end-effector, the Newton Raphson module solves for the optimal control input, which helps predict the next step of the forward kinematics.}
    \label{fig:predicted_hasel}
\end{figure}
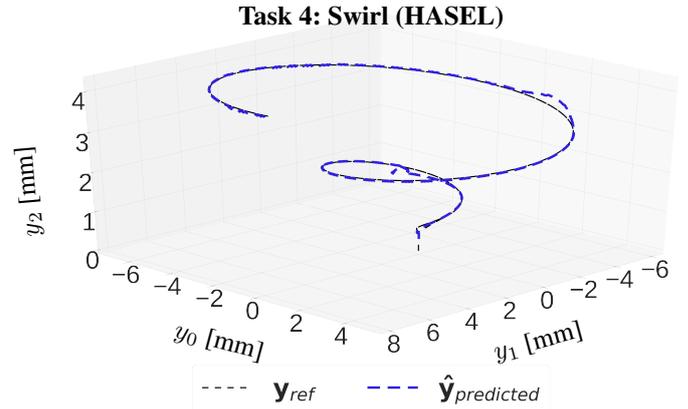

\subsubsection{HASEL-driven Path Following.} The tuning parameters used for the HASEL-driven platform were: $\mathbf{Q} = diag([1 \times 10^6, 2\times 10^6, 1\times 10^6])$, $\mathbf{\Lambda} = \mathbf{I}_{6 \times 6}$, $\epsilon = 1/130$, $N = 3$, $N_c = 1$, $s = 1\times 10^{-25}$, $b = 1 \times 10^{-5}$, $r = 4 \times 10^2$, $n_d = d_d = 2$. We achieve a root mean square error over one run of the \emph{Swirl} path of 0.05 mm between the firmware path in the controller and the reference path, as simulated in firmware. This error may increase when measuring the true end-effector positions using the motion capture due to marker noise and error between predicted and true paths (see Figure \ref{fig:forward_kino}(c). Figure \ref{fig:predicted_hasel} displays the path following results of the end-effector of the HASEL position that we monitor from the firmware through a serial cable. Therefore, the results on Figure \ref{fig:predicted_hasel} come from the embedded compute platform that actuates the HASEL actuators.

\begin{table*}
\small\sf 
\centering
\caption{Summary of model sizes and bandwidth compared to other state-of-the-art methods. \cite{grant2014cvx} are the developers of CVXGEN and \cite{boggs1995sequential} are the developers of SQP. Accuracy at the control task described in text.\label{T2}}
\begin{tabular}{ccccccccc}
\toprule
Related Work & Network Size  & Bandwidth & Solver & Feedback & Software & Computer \\
& $n_{size}$ [parameters]& [Hz] & (Y/N) & & \\
\midrule
This work (HASEL-driven) & 233 & 130 & Newton- & Y & C & Teensy 3.6 \\
 & &  & Raphson & & & \\
This work (tendon-driven) & 435 & 120 & Newton- & Y & Python/C & PC/ESP32$^a$\\
& & & Raphson & & &  \\
\cite{hyatt2019model} & 3.4 million & 20 & CVXGEN & Y &  MATLAB  & PC\\
\cite{gillespie2018learning} & up to 80,000 & 30 & CVXGEN & Y & MATLAB & PC  \\
\cite{thuruthel2019soft} & 1295 (worst case) & 50 & SQP & N & MATLAB & PC\\
\bottomrule
\end{tabular}\\
$^a$ Results from Figure \ref{fig:path_tracking} were collected from the controller running on a PC using the Python prototype.
\end{table*}

Table \ref{T2} summarizes the results from this work as compared to equivalent soft actuators and control algorithms that use neural network models. We compare the neural network size, frequency of the control loop in the experimental setup, the solver, whether the controller is a feedback controller and the software platform that the authors use. We do not consider path tracking accuracy toward this comparison due to the diversity of soft actuator designs in this body of literature and varying proneness to noisy responses of the actuator. However, the root mean squared error in this controller as measured by a motion capture system in the tendon-based platform does not exceed 2 $mm$, being this error larger in the \emph{Diagonal} task. For the HASEL-based platform, the root mean squared error as observed from the firmware predictions does not exceed 1 $mm$ (root mean square 0.05 $mm$, 0.03 $mm$ average absolute error, as simulated in the MCU firmware). In \cite{bruder2019modeling} the best case average error is 1.19 $cm$. In \cite{hyatt2019model} the best case steady state error in radians is 0.010 radians (0.57$^{\circ}$) for the control task, which is not path tracking control, similar to \cite{gillespie2018learning}, which uses model predictive control (MPC) to follow specific joint angle error was less than 12$^{\circ}$ and less than 2 degrees per second for angular velocity. For \cite{thuruthel2019soft}, the best case mean error is 3.8 cm and the standard deviation is 2.7 cm, which indicates low precision. We provide supplementary material on the untethering of the tendon-based platform. \footnote{\href{https://youtu.be/I17_4AV8vFg}{Link to untethered tendon-based demo.}}

\begin{table}
\small\sf 
\centering
\caption{Controller results summary.\label{T3}}.
\begin{tabular}{cccc}
\toprule
Mean & Max. Norm. & Max. Flash & Max. Bandwidth \\ 
Power & RMSE$^*$ & &\\ 
\midrule
45.35 mW & 5\% & 6.4\% & 130 Hz\\
\bottomrule
\end{tabular}\\
$^*$Maximum normalized root mean squared error (RMSE) is the maximum RMSE across all tasks divided over the maximum range in the task space (i.e. $\frac{max(\varepsilon_{RMSE})}{max(\Delta y)}$).
\end{table}

\subsection{Profiling}
\label{sec:profiling}

We profile this control method running in the MCU's firmware in terms of both compute time and power and energy consumption.

\subsubsection{Compute Time.} The change in solution time with respect to the prediction window follows a linear trend. We find that if the computation at $N = 1$ takes $\Delta t_0$ times in millisecons, the subsequent computations at $N > 1$ follows a relation of $\Delta t(N) = 0.118 N + \Delta t_0$. For the case of the tendon-based platform, the computation at $N = 1$ takes less than 5 $ms$ per control loop, for the case of the HASEL-based platform, the computation takes 7 $ms$ per control loop at $N = 1$.

\subsubsection{Flash Memory occupancy.}  When running in the Teensy 3.6 platform, we achieve a flash memory occupancy of 6.4\% of flash memory occupied (where only 67.3 $kB$ were used) and 2.4\% of RAM occupied (where only 6.3 $kB$ bytes are used). For the ESP32 the memory occupancy does not exceed 5.6\% of the available flash memory.

\subsubsection{Power Consumption.} We monitor the voltage and current using the power monitor for a period of one minute using a smaller (233 parameters) and a larger (2663 parameter) neural network for forward kinematic prediction under two configurations: idle (only bootstrapping the controller but no processing in a loop) and active (running the complete control algorithm). For the smaller neural network model, the average power is 261.72 $mW$ for active and 263.33 $mW$ for idle. For the larger model, the average power is 308.06 $mW$ when the controller is active and 262.33 $mW$ in idle mode, yielding an increase of 45.35 $mW$ from running the controller code. Considering a 1000 $mAh$ battery, the battery life for the smaller model is above 14 hours and for the larger model, it is 12.00 hours while the controller is active and 14.09 hours for idle. Based on the energy consumption curve, we model the energy consumption of this control method as a function of the neural network parameters ($n_{size}$) as Equation \ref{eq:model_energy_consumed}, which is an approximation that comes from the data collected from the power monitor based on regression from trendlines.
\vspace{-5pt}
\begin{equation}
    \label{eq:model_energy_consumed}
    \begin{aligned}
    & E(t, n_{size}) [\mu Ah] \approx (0.0008 n_{size} + 1.3) t-17.104 
    \end{aligned}
\end{equation}

\vspace{-15pt}

\section{Discussion}
\label{sec:discussion}

We achieve effective path tracking of the end-effector of two robotic platforms relying on non-linear soft actuators and sensors, and low-power MCUs. Transferring this method across platforms requires relatively minimal efforts: given data of the forward kinematics, we train a neural network to predict the output transition between discrete time step $k$ and $k+1$. Even though the accuracy of this control method compares favorably with state-of-the-art soft robotic model predictive controllers with learned models, a 1:1 comparison with this literature would be unfair due to the small-scale nature of the systems we test and the difference between the designs of the soft actuators, which may cause the noisier, lower-precision, responses we find in the surveyed literature.

The mass-spring-damper system is a suitable baseline to validate this approach because it approximates the behavior of the soft actuator without taking into account the nonlinearities of the soft composite material. This controller successfully approximates a PID controller on the mass-spring-damper (one dimensional) task with robust tuning. The steady-state error for the PID controller is lower (0mm), while this control approach a maximum steady-state error of 5.7mm on a simulated task space of 2m (which normalizes to ($\frac{0.0057m}{2m}*100 =$0.29\%). This error is within an acceptable range considering that the controller is based on a model learned from a system that does not have a first-principled model available. The former is an advantage of the controller, which makes it suitable for highly nonlinear systems made of soft, composite materials. 

This controller is also compatible with neural network models generated using Tensorflow Lite. The constant used for automatic differentiation may approximate to the inverse of the bandwidth in which the control loops operate. The gradients are applicable regardless of the number of inputs and outputs of the controller, which makes the setup more generalizable when presented with a novel soft actuating platform. 

In the tendon-based platform, the soft material is clamped on one side and unsupported against gravity on the other side, where only two actuators generate the movements within the task space. The HASEL-based platform is fixed on a plane that supports it against gravity, where the six actuators' task is to move the plane up and down. The two physical systems in this work present inherently different mechanical challenges. This explains the differences in the path following task results. We expect higher errors from the HASEL-based platform when we measure the path following from the motion capture system. 
 
 Results from this controller are also repeatable and high-precision regardless of the experimental and simulated platform and also noise coming from sensor signals or actuation. The choice of Newton-Raphson over the Jacobi method is justified by the guarantee of convergence of Newton-Raphson and the number of iterations that it takes to converge. After three iterations, the algorithm does not further improve its error, which is within an absolute tolerance of $10^{-4}$ in the solver.

The speed in which the reference path evolves affects the accuracy of the controller. This is not correlated with the bandwidth of the controller but on the inherent speed in which the end-effector needs to follow a trajectory. We report the bandwidth of the controller compared to other state-of-the-art model-free controllers in Table \ref{T2}. 

Regarding the recursive neural network model, there is a well-known drift in forward predictions as $N$ increases, which is why this controller works best for lower values of $N$, which also guarantees higher-bandwith operations. NARMA-L2 \cite{narendra1991learning} is an appropriate training method for shallow neural networks that also keeps the number of parameters low. However, these auto-regressive adaptive models require information on the input--output relationship of the system at all times past system identification to make these models adaptive and avoid recursivity, this makes a soft robot potentially reliant on the motion capture data under a NARMA-L2 control regime. The work presented in this paper aims to create control methods that allow for the soft robot to use its own proprioception to make decisions for full untethering based on internal networks of afferent sensors.

The limitations to this approach are: 1) the data collected often needs to be downsampled in order to match the frequency in which the controller is anticipated to follow; otherwise the forward kinematics predictions will not be accurate during experimental time; 2) the usage of heap memory as opposed to preallocating data structures in the stack to accelerate the computation; 3) the relationship between model size and maximum bandwidth that the controller may run at, which may be overcome through hardware accelerating techniques and model compression in order to have larger parameter-size neural networks do predictions in tasks that require more model complexity.

In the future, we envision smart composites to include large numbers of sensor/actuator systems such as described here. Here, co-locating computation and control will dramatically facilitate the routing of information in and out of the composite as all analog information can remain local to the sensors and actuators, requiring only a minimum of high-level, digital control information to be exchanged within different parts of the composite and outside controllers. 

\vspace{-6pt}
\section{Conclusion}

We use shallow, floating-point neural networks combined with numerical methods to achieve a lightweight, platform-independent closed-loop onboard control approach that is tractable with minimal computation that can be co-located with sensing and actuation. We provide tools to leverage these results for other soft actuator systems, only requiring a dataset on the inputs and outputs of the system and sensor signals. We show that this control technique generalizes to different soft actuators with different sensing technologies, computing platforms, and nonlinear properties with diverse challenges and nonlinearities that would otherwise affect the controller's performance. It also generalizes to canonical linear control problems, such as the mass-spring-damper system. Future work includes optimizing the neural network, control technique, and specialized linear algebra operations for specific target platforms, that is, using a set of libraries that will enable vectorization, usage of the digital signal processing (DSP) or field-programmable gate arrays (FPGA), and other powerful compute components already present at the low-power, low-cost platforms for soft actuators. 

By co-locating prediction and control with sensing and actuation, we are making steps towards enabling a new class of intelligent composite and paving the way for systems with large numbers of soft actuators working in parallel, not requiring a central computing system for their untethered operations nor an external observer that informs it of its state. However, the natural step that follows is the full integration of the embedded compute device into the composite material itself for such materials to be fully realized.

\vspace{-6pt}
\begin{acks}
We thank Prof. Hayes from CAIRO Lab for facilitating the motion capture equipment, Dr. Soloway, the Air Force Office for Research (Grant No. 83875-11094) and Hoang Truong for facilitating the power monitoring device.
\end{acks}
\vspace{-4pt}

%%%%% FOR SUBMISSION:

%\bibliography{references.bib}

\end{document}